\documentclass[lettersize,journal]{IEEEtran}
\usepackage{amsmath,amsfonts}
\usepackage{algorithmic}
\usepackage{algorithm}
\usepackage{array}
\usepackage[caption=false,font=normalsize,labelfont=sf,textfont=sf]{subfig}
\usepackage{textcomp}
\usepackage{stfloats}
\usepackage{url}
\usepackage{verbatim}
\usepackage{graphicx}
\usepackage{cite}
\usepackage{tikz}
\usepackage{comment}
\usepackage{amsmath,amssymb} 
\usepackage{color}
\usepackage{graphicx}
\usepackage{amsmath}
\usepackage{amssymb}
\usepackage{booktabs}
\usepackage{graphicx}
\usepackage{xcolor}
\usepackage{colortbl}
\usepackage{nicematrix}

\usepackage{textcomp}
\usepackage{stfloats}
\usepackage{url}
\usepackage{verbatim}
\usepackage{graphicx}
\usepackage{cite}
\usepackage{tikz}
\usepackage{comment}
\usepackage{amsmath,amssymb} 
\usepackage{color}
\usepackage{tabularx,lipsum}

\usepackage{wasysym}
\usepackage{graphics} 
\usepackage{epsfig} 
\usepackage{mathptmx} 
\usepackage{color} 
\usepackage{tabularx}
\usepackage{makecell}
\usepackage{multirow}
\usepackage{mwe,lipsum}
\usepackage{array,multirow}
\usepackage{float}
\usepackage{eucal}
\usepackage{comment}
\usepackage{pifont}

\usepackage{arydshln}

\usepackage{wasysym}
\usepackage{graphics} 
\usepackage{epsfig} 
\usepackage{mathptmx} 
\usepackage{color} 
\usepackage{tabularx}
\usepackage{makecell}
\usepackage{multicol}
\usepackage{mwe,lipsum}
\usepackage{array,multirow}
\usepackage{float}
\usepackage{eucal}
\usepackage{xcolor}

\newcommand{\bl}[1]{{\textcolor{black}{\textbf{#1}}}}

\begin{document}

\title{HiDAnet: RGB-D Salient Object Detection via Hierarchical Depth Awareness}

\author{Zongwei Wu, Guillaume Allibert, Fabrice Meriaudeau, Chao Ma, and C\'edric Demonceaux 
\thanks{Z. Wu, F. Meriaudeau, and C. Demonceaux are with ImViA, Université Bourgogne Franche-Comté, Dijon, France (e-mail: \{zongwei\_wu@etu.; fabrice.meriaudeau@; cedric.demonceaux@\}u-bourgogne.fr )}
\thanks{G. Allibert is with Universit\'e  C\^ote d’Azur, CNRS, I3S, Nice, France (e-mail: allibert@i3s.unice.fr)}
\thanks{Z. Wu and C. Ma are with MOE Key Lab of Artificial Intelligence, AI Institute, Shanghai Jiao Tong University, Shanghai, China (chaoma@sjtu.edu.cn)}
}

\markboth{Journal of \LaTeX\ Class Files,~Vol.~14, No.~8, August~2021}%
{Shell \MakeLowercase{\textit{et al.}}: A Sample Article Using IEEEtran.cls for IEEE Journals}


\maketitle

\begin{abstract}
RGB-D saliency detection aims to fuse multi-modal cues to accurately localize salient regions. Existing works often adopt attention modules for feature modeling, with few methods explicitly leveraging fine-grained details to merge with semantic cues. Thus, despite the auxiliary depth information, it is still challenging for existing models to distinguish objects with similar appearances but at distinct camera distances. In this paper, from a new perspective, we propose a novel Hierarchical Depth Awareness network (HiDAnet) for RGB-D saliency detection. Our motivation comes from the observation that the multi-granularity properties of geometric priors correlate well with the neural network hierarchies. To realize multi-modal and multi-level fusion, we first use a granularity-based attention scheme to strengthen the discriminatory power of RGB and depth features separately. Then we introduce a unified cross dual-attention module for multi-modal and multi-level fusion in a coarse-to-fine manner. The encoded multi-modal features are gradually aggregated into a shared decoder. Further, we exploit a multi-scale loss to take full advantage of the hierarchical information. Extensive experiments on challenging benchmark datasets demonstrate that our HiDAnet performs favorably over the state-of-the-art methods by large margins. 
\end{abstract}


\section{Introduction}

Salient object detection (SOD) aims to find the most prominent region inside an image that visually attracts human attention. Conventional SOD approaches only take color images as inputs. With deep learning models, RGB SOD has achieved significant success \cite{zhang2017learningsingle,deng2018r3net,liu2019simple,wu2019cascaded,zhao2019egnet}. However, these models may result in unsatisfactory performance when dealing with complex scenes, e.g., low-contrast light or object occlusion.

Recent advanced RGB-D sensors provide accessibility to depth maps at a low cost. The complementary geometric cues can contribute to scene understanding. In the literature, two main designs have been widely exploited, i.e., single-streaming schemes that combine RGB-D images from the input side \cite{zhao2020single,fu2020jldcf,zhang2020uc} and multi-streaming network that extracts multi-modal features separately and combines them at semantic levels \cite{fan2020bbs,ji2021calibrated,CDINet,cascaded_cmi,zhouiccvspnet,liu2021TriTrans,sun2021deep,Zhang2021DFMNet}. Existing networks often directly extract semantic features through the deep network, with few methods fully explore the rich geometric priors provided by the depth map.

\begin{figure}[t]
\centering
\includegraphics[width=\linewidth,keepaspectratio]{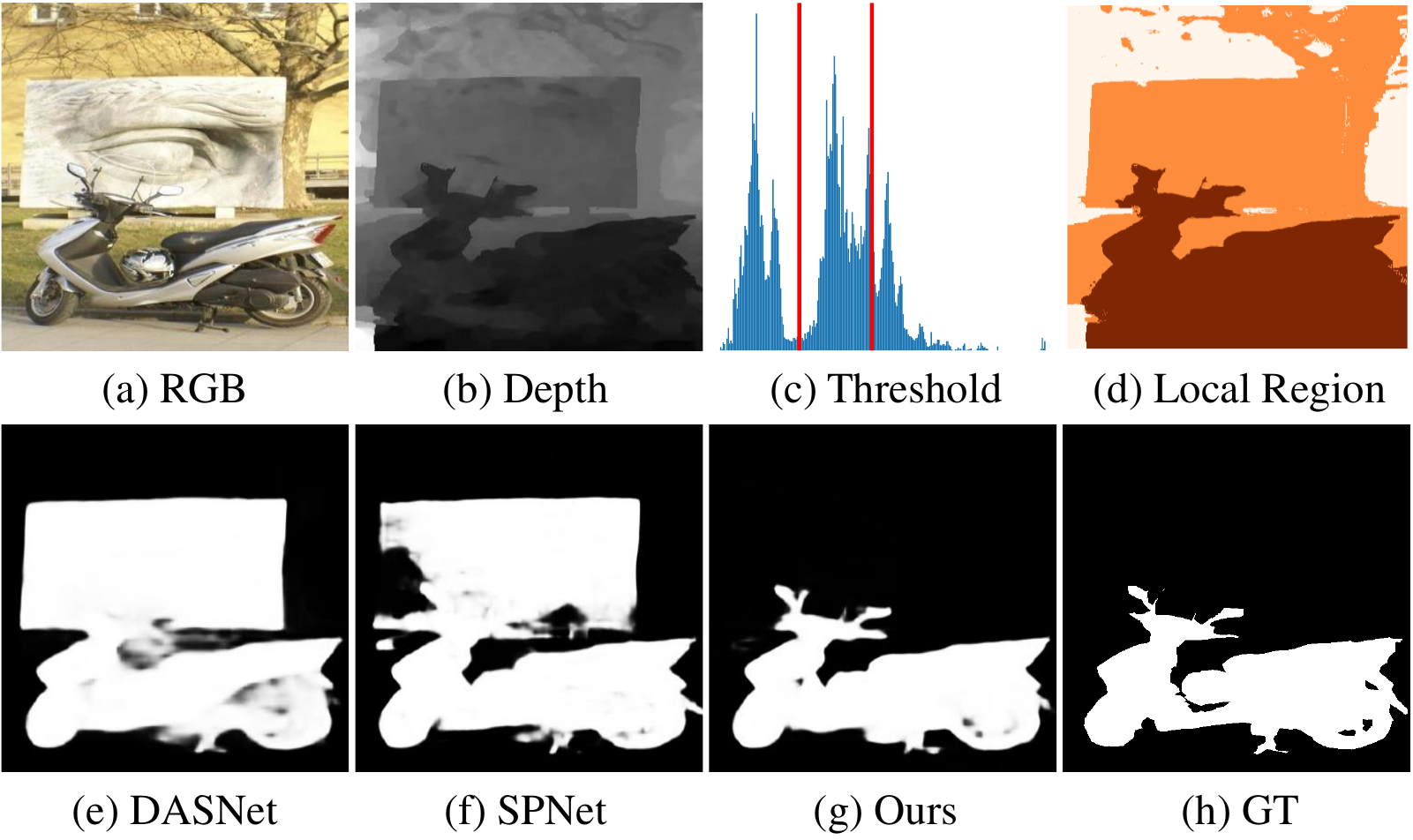}
\caption{\textbf{Motivation} of our hierarchical depth awareness. (a) and (b) are the paired RGB-D inputs. (c) and (d) represent Multi-Otsu thresholding on depth histogram and the generated Otsu regions, respectively. Our approach takes full advantage of depth priors to improve the feature discriminatory power and obtain the saliency mask (g). Compared to two state-of-the-art (SOTA) RGB-D models (e) and (f), our method favorably yields results (g) closer to the ground-truth mask (h). }
\label{fig:intro}
\end{figure}

Previous works on channel attention \cite{woo2018cbam,hu2018senet,He2016Residual,wang2020eca} have shown their effectiveness in emphasizing the attentive features among channels. A number of saliency detection works \cite{fan2020bbs,zhao2020depth,ji2021calibrated,cascaded_cmi} adopt channel attention to enhance multi-modal features. However, the first step of learning channel attention is to aggregate the spatial information of feature maps to construct a $1\times 1 \times C$ vector by using global average pooling, where $C$ is the number of channels. As a result, the foreground and background contribute equally to the output, which is not optimal to distinguish salient objects. Considering these issues, an intuitive motivation is to design local channel attention referring to depth priors in order to improve feature representation learning.  

As shown in Fig. \ref{fig:intro}, while dealing with complex scenes, current state-of-the-art (SOTA) RGB-D models \cite{zhao2020depth,zhouiccvspnet} fail to extract the salient region due to similar visual appearance between the foreground and background  (Fig. \ref{fig:intro}(f) and (g)). However, we observe that salient regions often share similar depth properties, i.e., a certain granularity of depth prior, that help to distinguish the salient objects from the background (Fig.\ref{fig:intro}(b) and (d)). Inspired by this observation, we develop a local feature enhancement scheme with granularity-based attention (GBA) to improve saliency detection. Specifically, we propose to first generate various local regions according to the granularity via Otsu thresholding \cite{otsu1979threshold,liao2001multiotsu}. These regions can be considered as distinct local spatial attention. Then for each region, we apply local channel attention to improve the feature discriminatory power. Fig. \ref{fig:intro}(c) and (d) illustrates such an example of the Otsu threshold values and granularity-aware masks, respectively. We show that our approach can better reason about salient regions (Fig. \ref{fig:intro}(g)) that are closer to the ground truth (Fig. \ref{fig:intro}(h)).

We further introduce a cross dual-attention module (CDA) to learn channel and spatial attention from auxiliary modalities to improve the current streaming. The enhanced features are hierarchically fused for final saliency map generation. Besides, the same cross-interaction scheme is embedded to articulate features between encoders and decoders through a U-Net-like \cite{ronneberger2015unet} architecture. We attentively mirror the multi-scale encoder features to preserve valuable geometric priors within each decoder. The encoded features are gradually fused to a shared decoder. Finally, we use a multi-scale loss on top of outputs from each decoder to optimize the saliency map. Concretely, our contributions are summarized as follows:

\begin{itemize}
\setlength{\itemsep}{1pt}
\setlength{\parsep}{0pt}
\setlength{\parskip}{0pt}
    \item We propose a novel granularity-based attention scheme that attends to fine-grained details in order to strengthen the feature discriminability of each modality. 
    \item We design a new multi-modal and multi-level fusion scheme with a multi-scale loss to take full advantage of the network hierarchy. 
    \item We extensively validate our HiDAnet on large-scale challenging benchmarks. Our approach performs favorably over SOTA models with large margins. 
    
\end{itemize}

\section{Related Work}

There are extensive surveys \cite{deepsodsurvey,shen2019survey,zhao2019odsurvey,cong2018sodsurvey,borji2019sodsurvey,zhou2021rgbdsurvey} of salient object detection in the literature. In this section, we briefly review related RGB-D saliency detection as follows: 

\noindent\textbf{Multi-Modal Fusion. } The auxiliary depth map provides extra geometric clues in addition to visual appearance. To efficiently merge both modalities, several fusion methods have been proposed. A number of works \cite{zhao2020single,fu2020jldcf,zhang2020uc,fu2021siamese,zhang2021uncertainty,chen20223} directly concatenate the depth map with RGB images from the input side through a single-stream network.  On the one hand, JLDCF and its successor \cite{fu2020jldcf,fu2021siamese} explore the siamese design for saliency detection by concatenating RGB and depth images in an additional dimension with a joint learning scheme. DANet \cite{zhao2020single} forms a four-channel input and enhances the extracted features with a dual-attention mechanism learned from depth. \cite{zhang2020uc,zhang2021uncertainty} propose the stochastic framework to analyze the uncertainty during human labeling and model the distribution of the saliency output. Different from previous works, \cite{chen2021rgb,chen20223} attempt to address RGB-D SOD from the 3D point of view with a 3D convolutional neural network. The recent \cite{zhou2022mvsalnet} leverages the depth cues to mimicks multi-view images and then fuse them to form the final output.

On the other side, multi-stream models \cite{fan2020bbs,ji2021calibrated,CDINet,cascaded_cmi,zhouiccvspnet,liu2021TriTrans,sun2021deep,Zhang2021DFMNet,lee2022spsn} have achieved leading performances in RGB-D SOD.  These models adopt two parallel encoders on different modalities, and the features are fused through different strategies. Several works \cite{zhu2019pdnet,fan2020bbs,zhai2021bifurcated} firstly enhance the depth features before fusing with RGB features. It is worth noting that a portion of the depth maps in existing saliency datasets are not of satisfactory quality. As discussed in \cite{fu2020jldcf,fan2019rethinkingd3,wu2022robust,cheng2022depth}, the depth may contain measurement or estimation bias. Thus, DCF \cite{ji2021calibrated} designs a calibration module to improve the depth quality. \cite{cheng2022depth,wu2022robust,huang2022discriminative,lee2022spsn} propose a layer-wise attention to model the geometric contribution with respect to the network depth. \cite{cheng2022depth} explores an additional backbone to learn the weighting scalar purely from depth. \cite{wu2022robust} analyzes the similarity between RGB and depth features to regular the depth contribution. Sharing the same motivation, \cite{lee2022spsn} computes the reliability of each modality at each stage and then merges them through their reliability. Instead of learning the weighting scalar, \cite{huang2022discriminative} generates the weighting maps at each scale to calibrate the feature response. Similarly, \cite{zhang2021bilateral} leverages bilateral attention to improve foreground-background features separately. Unlike these works, we first divide the feature map into several local regions with the help of depth granularity. The feature maps are further calibrated with different local attention to improve the feature discriminability. Compared to \cite{huang2022discriminative,zhang2021bilateral}, our fined-grained details are 
 statically computed by maximizing the inter-class distance without learning parameters, leading to more reasonable and stable locally-calibrated areas. 

There exist other works which only extract features from RGB input while the depth map only serves as supervision \cite{piao2020a2dele,ji2020accurate,zhao2020depth}. In this context, \cite{wu2021modality,jin2021cdnet} propose to leverage the pseudo-depth to guide the RGB learning. A2dele \cite{piao2020a2dele} further formulates depth supervision as a knowledge transfer problem. CoNet \cite{ji2020accurate} and DASnet \cite{zhao2020depth} propose a multi-task learning framework with an additional depth head together with the saliency branch. However, we argue that these methods cannot fully leverage the multi-modal cues during feature extraction. Instead, we propose a cross-interaction scheme to take full advantage of cross-modal cues. We benefit from the auxiliary modality to alleviate errors in the feature modeling (depth to RGB, and RGB to depth).

\noindent\textbf{Multi-Level Fusion.} U-Net with skip connections \cite{ronneberger2015unet} has shown its effectiveness in pixel-level segmentation tasks. Several RGB-D SOD models \cite{pang2020hierarchical,zhouiccvspnet,liu2021TriTrans,CDINet} equip this design for clearer boundary generation. \cite{pang2020hierarchical} adopts the feature-wise addition. \cite{zhouiccvspnet,liu2021TriTrans} concatenate the encoder features with the decoder. \cite{CDINet} designs a dense connection between high-level features and the decoder. In this work, we exploit the contribution of attention modules for skip connections applied to SOD. It is worth mentioning the success of skip connections can be mainly attributed to aggregation between the semantic features provided by the contracting path and fine-grained features from the expansion path. From a new perspective, we consider the encoder-decoder features as multi-modal features, and a unified cross-fusion scheme is applied to boost the performance. 

\noindent\textbf{Attention for Feature Enhancement.} Attention methods such as transformer \cite{vaswani2017attention}, CBAM \cite{woo2018cbam}, SEnet \cite{hu2018senet}, DA \cite{fu2019dual}, and ECA \cite{wang2020eca} have demonstrated their success in other vision tasks. A number of RGB-D saliency models also equip attention modules to extract attentive features from different modalities. VST \cite{liu2021vst} and TriTrans \cite{liu2021TriTrans} adopt transformer \cite{vaswani2017attention} for saliency detection. \cite{zhao2021rgb,zhao2020depth,wang2022learning} apply the SE module to compute modality-specific attention for feature calibration. Similarly, CDInet \cite{CDINet} designs a depth-induced channel attention to enhance RGB features. From another perspective, \cite{zhang2022c} deeply explores the spatial attention at different scales with the help of decoupled dynamic convolution. Sharing the same motivation, DFMnet \cite{Zhang2021DFMNet} adopts a depth holistic attention on top of features with different resolutions. More recently, several works leverages both spatial and channel attention to jointly improve the feature representation. For example, BBSnet \cite{fan2020bbs} applies the CBAM \cite{woo2018cbam} on the depth map to improve the depth quality before fusion. \cite{wen2021dynamic} further improves the CBAM by highlighting spatial features. Sharing the same motivation, CMINet \cite{cascaded_cmi} applies the DA \cite{fu2019dual} on to lately merge RGB-D features. Different from previous works with bi-directional cross-modal attention, HAINet \cite{li2021hierarchical} explores the purified depth to improve the RGB features in turn.

Despite the proven effectiveness, previous channel attention schemes do not fully benefit from the geometric priors. For example, the same attention can be applied to both foreground and background. The rich geometric priors in the input depth map have rarely been discovered, which limits the performance of RGB-D saliency detection. DSA2F \cite{sun2021deep} introduces a depth-sensitive module with the help of the depth histogram. However, it computes the depth region with a fixed threshold for each input image and the attention scores are simply computed by a $Conv_{1\times 1}$. In contrast, we propose to dynamically generate multi-granularity regions with the multi-Otsu method \cite{otsu1979threshold,liao2001multiotsu}. The fine-grained details are further integrated with channel attention to enhance the feature discriminability for sharper edge generation.

\section{Method}

\begin{figure*}[t]
\centering
\includegraphics[width=\linewidth,keepaspectratio]{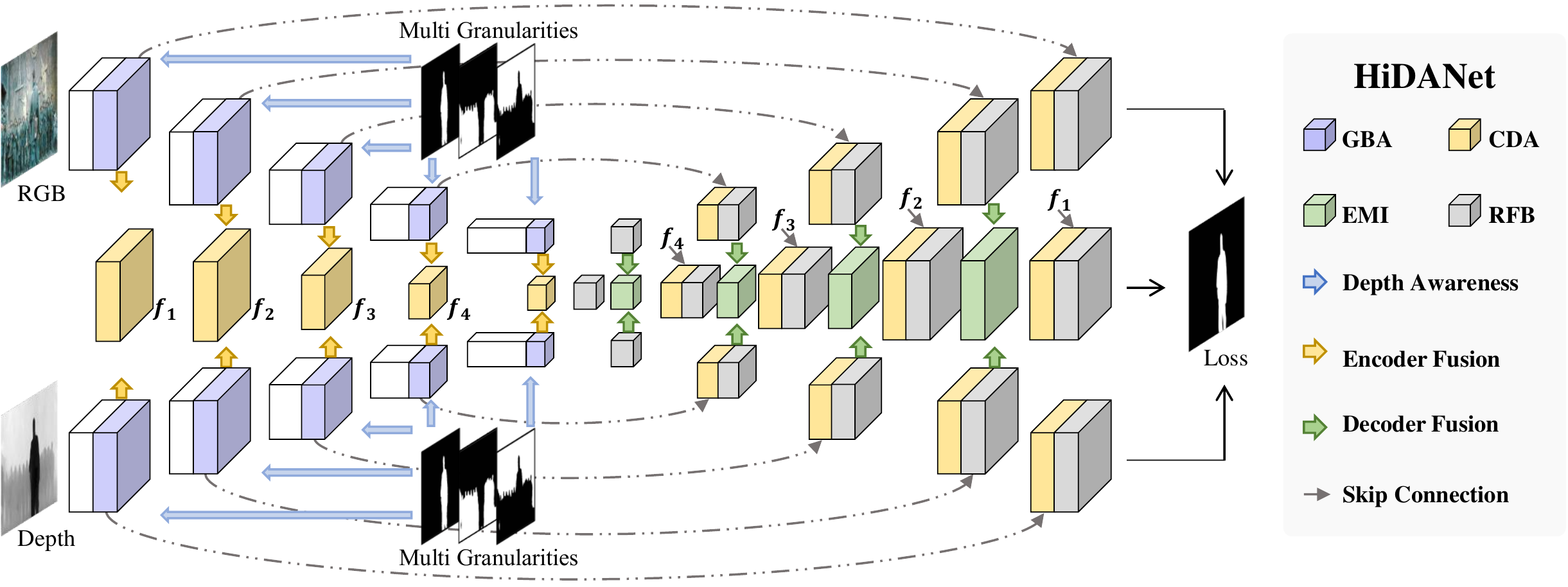}
\caption{The overall architecture of our HiDAnet with U-Net-like design. It consists of granularity-based attention (GBA Section \ref{gba}), cross dual-attention module (CDA Section \ref{cda}), and efficient multi-input fusion (EMI Section \ref{emi}). RFB is the receptive field block from \cite{liu2018rfb} for accurate object detection. White blocks denote the network backbone. Our granularity-based attention strengthens the discriminatory power of RGB and depth features separately. Our cross dual-attention module takes advantage of cross-domain cues to attentively realize multi-modal and multi-level fusion in a coarse-to-fine manner. Our efficient fusion scheme effectively models the shared information from each modality. The shared features are further improved with the skip connections for final saliency map generation. Best viewed in color.}
\label{fig:hidanet}
\end{figure*}

Fig. \ref{fig:hidanet} presents the overall framework of our proposed HiDAnet. Note that the Otsu masks are generated from the depth map during the pre-processing. Firstly, RGB and depth maps are fed into two parallel encoders for feature extraction. For each individual encoder (RGB/Depth), we propose a granularity-based module (GBA) with the help of input Otsu masks to enhance the discriminatory power, e.g., foreground and background. This module is naturally embedded into different levels of the encoder to correlate with the network hierarchies. With the enhanced features, we propose a unified fusion mechanism (CDA) for multi-modal and multi-level fusion. It enables a cross-domain interaction with both channel and spatial attention to learn the informative shared features in a coarse-to-fine manner. These features are later gradually aggregated into the shared decoder through the efficient multi-input fusion module (EMI). Lastly, we exploit a multi-level loss to take full advantage of the network hierarchies. Details of each component are presented in the following sections.

\subsection{Feature Extraction with Granularity-Based Attention}
\label{gba}

\begin{figure*}[t]
\centering
\includegraphics[width=\linewidth,keepaspectratio]{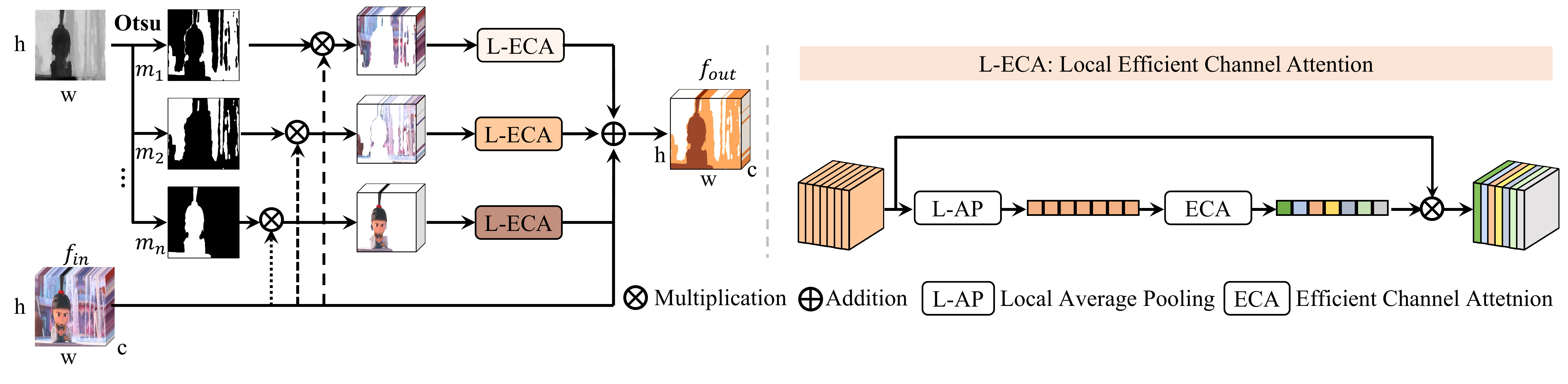}
\caption{Diagram of the \textbf{granularity-based attention}. The depth awareness is encoded via Local Efficient Channel Attention (L-ECA). ECA is from \cite{wang2020eca}.}
\label{fig:grain}
\end{figure*}

We observe that the multi-granularity properties of geometric priors correlate well with the network hierarchies of saliency models. Inspired by this observation, we propose the granularity-based attention that aims to attentively combine the spatial attention mask with the conventional channel attention as shown in Fig. \ref{fig:grain}. For earlier layers, it strengthens the low-level representations to precisely localize the salient object with a sharp boundary. For deeper layers, it improves the semantic abstraction and contributes to the identification of salient objects regardless of appearance variations. 

Given the depth map $D$ with its histogram $H$, we dynamically generate the fine-grained details. According to the value/distance within the depth map, we use the Otsu algorithm \cite{otsu1979threshold} to discretize the histogram H into several different regions. 
In this work, we use the extended multi-Otsu \cite{liao2001multiotsu} to generate multiple thresholds. Assuming $T$ random thresholds $(d_1, d_2, ..., d_T)$ dividing the depth into $T+1$ parts. Let ($\sigma^2_{i},w_{i}$) be the variance and the pixels number of region $i$ ($1 \leq i \leq T+1$). The optimal values $\{d_1^*, d_2^*, ..., d_{T}^*\}$ are chosen by maximizing the inter-class variance:
\begin{equation}
   \{d_{1}^*, d_2^*, ..., d_T^*\} = argmax\{\sigma_{w}^{2}(d_1, d_2, ..., d_T)\},
   \label{multiotsu}
\end{equation}
where $\sigma_{w}^{2} = \sum_{i=1}^{T+1} w_{i}\sigma^2_{i}$. To reduce the computational cost, we only generate the Otsu regions once during pre-processing and further resize them to fit the resolution of feature maps from different scales.


\begin{figure*}[t]
\centering
\includegraphics[width=\linewidth,keepaspectratio]{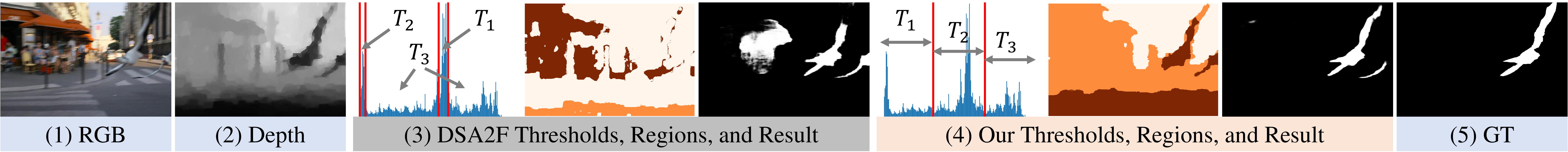}
\caption{\textbf{Visual comparison} with concurrent DSA2F \cite{sun2021deep}. Our method maximizes the inter-class variance, leading to more accurate masks compared to DSA2F. We further explore the granularity cues via channel attention, yielding results closer to the ground truth (5).}
\label{fig:diff}
\end{figure*}

For the $i^{th}$ region $m_i$, ($1\leq i \leq T+1, i\in \mathbb{N^*}$), we mask out the feature map $f_{in}$ with element-wise multiplication to suppress the inactive area through $f_{in} \otimes m_i$. Then, the channel attention is applied to improve the feature representation with local awareness. Compared to the vanilla channel attention \cite{hu2018senet,wang2020eca}, we replace the global average pooling with the local average pooling that attends to the local details referring to geometric priors. Finally, the locally enhanced features are aggregated by a residual connection for the final output generation $f_{out}$. The overall process can be formulated as:
\begin{equation}
\begin{split}
&L-ECA(x) = \sigma(Conv_{1d}(L-AP(x))) \otimes x, \\
&f_{out} = \sum_{i=1}^{T+1} L-ECA(f_{in} \otimes m_i) + f_{in},
\end{split}
\end{equation}
where $\sigma(\cdot)$ is the Sigmoid activation, $\otimes$ is the element-wise multiplication, and $L-AP$ denotes the local average pooling on each masked region.  We provide more details on the differences between the proposed granularity-based attention and traditional channel attention in the ablation study Section \ref{abla:gba} Tab. \ref{tab:reb}.

\noindent\textbf{Remarks.} Several previous works have proposed to explore depth prior in various manners such as the contrast in CPFP \cite{zhao2019contrast}, the edge in CoNet \cite{ji2020accurate}, or the histogram in DSA2F \cite{sun2021deep}. Our approach resembles the DSA2F that both methods belong to threshold-based segmentation frameworks. However, one main difference is that we dynamically generate optimized masks with the Ostu algorithm, while DSA2F applies fixed thresholds on the $T+1$ largest depth distribution modes that cannot adapt to different scenarios without handcraft adjusting. Fig. \ref{fig:diff} illustrates the difference in the thresholds and regions. We observe that our approach computes more discriminative regions, yielding a more effective and robust manner to explore the depth prior.  Moreover, since the Otsu algorithm optimizes the thresholds by maximizing inter-class variance, our generated masks are more robust to the depth noise compared to the concurrent work. Additionally, we leverage the granularity with channel attention, while DSA2F simply uses a $Conv_{1\times 1}$ for local awareness. As shown in Fig. \ref{fig:diff}, by integrating the fine-grained details into the channel attention, we can reason about more accurate saliency regions closer to the ground truth. The quantitative comparison with \cite{ji2020accurate,zhao2019contrast,sun2021deep} can be found in Section \ref{quantitative} Tab. \ref{tab:quant}.  Our superior performance proves that we can better model the depth priors.


\subsection{Encoder Fusion with Cross Dual-Attention Module}
\label{cda}

Previous studies \cite{fan2019rethinkingd3,piao2020a2dele,zhao2020depth} have affirmed the effectiveness of learning from two heterogeneous modalities for RGB-D SOD. Color images provide rich information in visual appearance while depth maps contain more spatial priors. Both modalities contribute to modulating homogeneous semantic information. Therefore, the objective of multi-modal learning is to efficiently fuse features with diverse information from different modalities. Similar to multi-modal features, multi-level features also contain both heterogeneous and homogeneous information: high-level features are richer in abstract semantic cues while low-level features are richer in fine-grained details. Thus, from a new perspective, we design a unified fusion scheme to make full use of cross-domain cues for both multi-modal and multi-level reasoning.

Assuming two paired multi-modal features $f_{x}$ and $f_{y}$. We firstly build a transformation $F_t$ to map the inputs $f_{x}, f_{y} \in \mathbb{R}^{C\times h \times w}$ to feature maps $f'_{x}, f'_{y} \in \mathbb{R}^{C'\times h \times w}$ with $C' = \frac{C}{2}$. Specifically, $F_t$ is the combination of a $1\times 1$ convolution which halves the channel size and a $3\times 3$ convolution which is expected to activate the edge response:

\begin{equation}
\begin{split}
    f'_{x} = F_t (f_x) = Conv_{3\times3}(Conv_{1\times1}(f_x)), \\
    f'_{y} = F_t (f_y) = Conv_{3\times3}(Conv_{1\times1}(f_y)).
\end{split}
\end{equation}

Once obtaining the lightweight representation, the next step is to aggregate features from different domains (RGB-D or encoder-decoder). We observe from Fig. \ref{fig:intro} that the fine-grained details, such as relative boundary, facilitate the identification of salient objects. Simultaneously, in case it is difficult to distinguish objects at the same distance on the depth map, e.g., when distinguishing the motorbike from the street, the visual appearance becomes more reliable. Inspired by this observation, we aim to use heterogeneous clues to compensate for the single-domain streaming.

\begin{figure}[t]
\centering
\includegraphics[width=\linewidth,keepaspectratio]{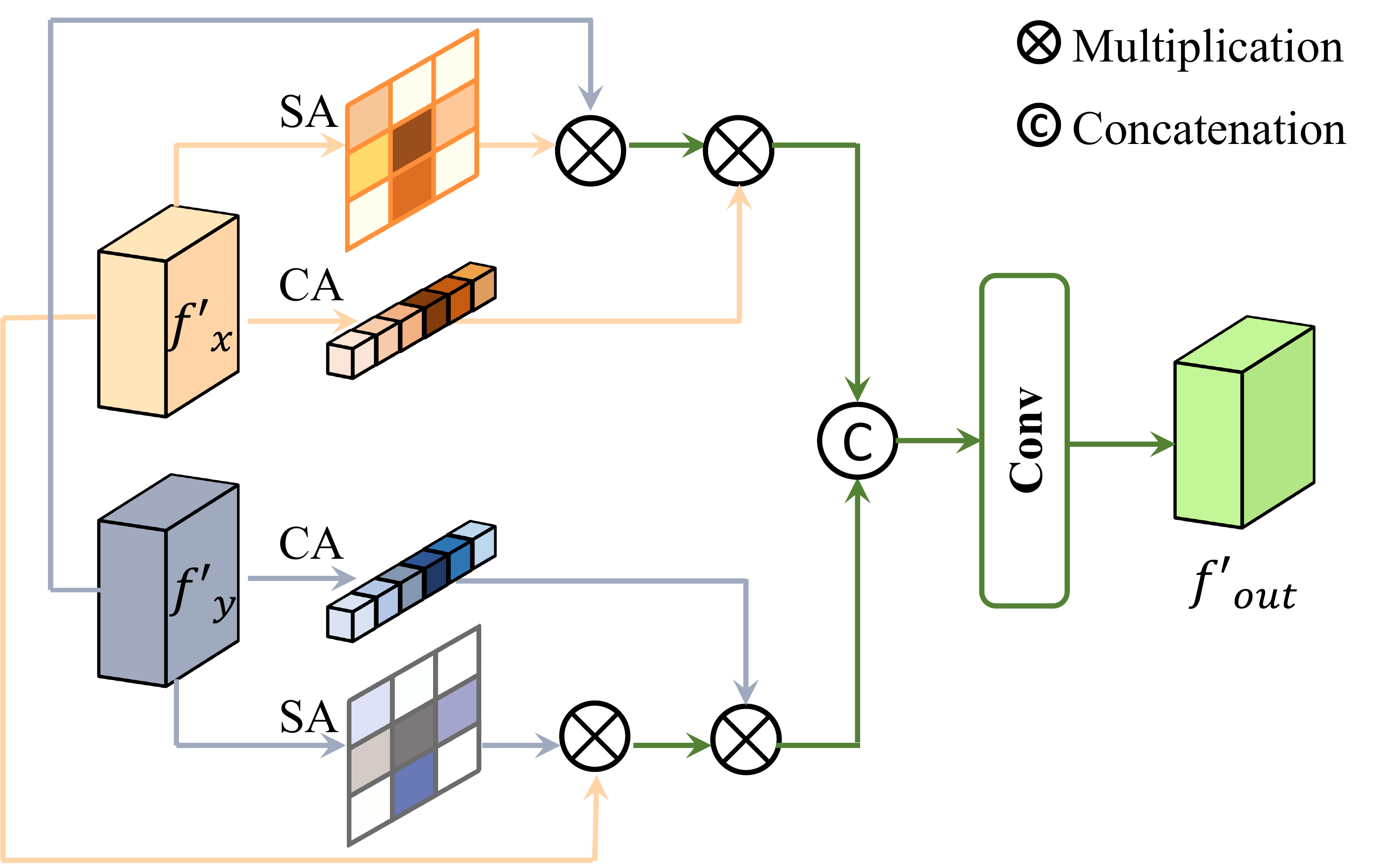}
\caption{The proposed multi-scale multi-level \textbf{encoder fusion} scheme with cross-domain supervision. Best viewed in color.}
\label{fig:fusion}
\end{figure}

To this end, we propose a cross dual-attention fusion scheme as shown in Fig. \ref{fig:fusion}. Specifically, from each input feature map, we learn the 1-D channel attention $M_c \in \mathbb{R}^{C'\times 1 \times 1}$ to determine \textit{what} information to be involved, and the 2-D spatial attention $M_s \in \mathbb{R}^{1\times h \times w}$  to determine \textit{which} part to focus. We formally have the operations:
\begin{equation}
\begin{split}
&M_c(f') = \sigma(MLP(GAP(f')) + MLP(GMP(f'))), \\
&M_s(f') = \sigma(Conv_{7\times7}(Concat(CAP(f'), CMP(f')))),
\end{split}
\end{equation}
where $\sigma(\cdot)$ is the Sigmoid activation, MLP is the multi-layer perceptron,  GAP and GMP are the global average and max pooling, respectively, and CAP and CMP are the average and max pooling across the channel, respectively. With the learned dual attention from separate feature maps, we enable a cross-domain interaction. In such a way, we can alleviate the ambiguities in the domain-specific features. Finally, the cross-enhanced features are fed into concatenation and convolution to form the shared representation $f'_{out}$. The overall process can be formulated as:
\begin{equation}
\begin{split}
&f^{enh}_{x} = M_s(f'_y) \otimes M_c(f'_y) \otimes f'_x, \\
&f^{enh}_{y} = M_s(f'_x) \otimes M_c(f'_x) \otimes f'_y, \\
&f'_{out} = Conv_{3\times 3} (Concat(f^{enh}_{x}, f^{enh}_{y})) ,
\end{split}
\end{equation}
where $\otimes$ denotes element-wise multiplication. For the shared encoder, starting from the second layer, once the multi-modal features are fused through cross attention, the output is further combined with the previous level output through a $Conv_{3\times 3}$.

\noindent\textbf{Remarks.} Our fusion design differs from concurrent works \cite{zhao2020depth,ji2021calibrated,zhang2020asymmetric,zhouiccvspnet} in several aspects: (\textbf{A}) We leverage both spatial and channel attention to aggregate multi-modal features, while \cite{zhao2020depth,ji2021calibrated} only focus on channels; (\textbf{B}) Different from ASTA \cite{zhang2020asymmetric}, our calibration is bi-directional (RGB to depth and depth to RGB), while ASTA is asymmetric which only leverages depth cues to improve RGB features. Hence, it does not tackle depth noise; (\textbf{C}) SPNet \cite{zhouiccvspnet} also adopts the symmetric fusion strategies. Our work differs from SPNet in that we fully explore the attention modules for feature fusion, while SPNet is built upon simple convolutions to combine features; (\textbf{D}) The fusion scheme can also be implemented by the CBAM \cite{woo2018cbam}. However, vanilla CBAM is modality-specific and cannot explore its relevance in cross-domain features. The ablation study in Section \ref{abla:hida} Tab. \ref{tab:ablacomp} shows the gain with the cross interaction.

\subsection{Decoder Aggregation with Efficient Multi-Input Fusion Module}
\label{emi}

\begin{figure}[t]
\centering
\includegraphics[width=\linewidth,keepaspectratio]{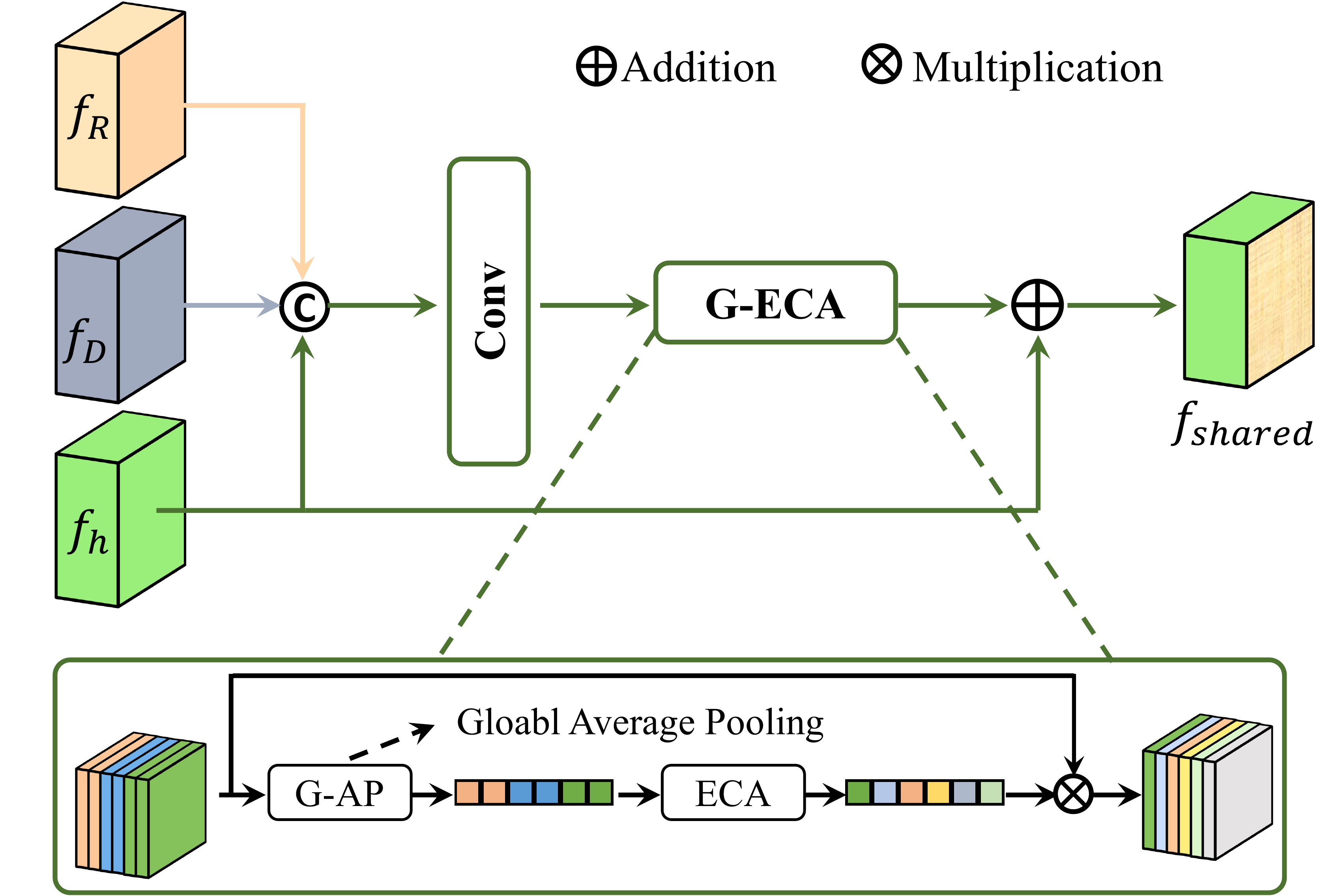}
\caption{The proposed efficient \textbf{decoder fusion} scheme for multi-type inputs. By fully exploiting the channel-wise dependencies, input features are attentively aggregated to generate the shared output. Best viewed in color.}
\label{fig:decoder}
\end{figure}

To aggregate the learned features from both RGB and depth decoders into the shared decoder, a simple concatenation may not be adaptive enough due to the tripled number of descriptors. Thus, we propose an efficient multi-input fusion strategy. Specifically, as shown in Fig. \ref{fig:decoder}, after the simple concatenation between different inputs (RGB $f_R$, depth $f_D$, and previous-level shared features $f_h$), we adopt the vanilla ECA \cite{wang2020eca} module (termed G-ECA with global pooling) to explore the inter-dependencies of different features. Thus, the most responded features are adaptively selected to form the shared decoder. A residual addition is adapted to reinforce the contribution of the previous level features. We have the overall process:
\begin{equation}
f_{shared} = G-ECA(Conv_{3\times 3}(Concat(f_{R}, f_{D}, f_{h}))) + f_{h}.
\end{equation}

The shared decoded features are then fed into our cross dual-attention scheme to realize the skip-connection between the shared encoder-decoder. 

\noindent\textbf{Remarks.} Our encoder fusion (CDA) and decoder fusion (EMI) are technically different. We observe that the spatial cues are gradually lost during encoding and become limited for decoders. This motivates us to apply both spatial and channel attention for the encoder fusion, while only using channel attention for the decoder fusion. 

\subsection{Optimization}

To take full advantage of the hierarchical information, we supervise multi-level outputs for both RGB, depth, and shared/fused branches. For outputs from each level, the predicted map is upsampled to form the same resolution mask as the ground truth. We adopt BCE loss $\mathcal{L}^{BCE}$ for pixel restriction and IoU loss $\mathcal{L}^{IoU}$ for global restriction \cite{qin2019basnet,wei2020f3net,zhao2020depth}. Therefore, we have the loss $\mathcal{L}_i$ for the $i^{th}$ level output:
\begin{equation}
   \mathcal{L}_i = \mathcal{L}_i^{BCE} + \mathcal{L}_i^{IoU}.
\end{equation}

 In total, we have five-level outputs (after each RFB in Fig. \ref{fig:hidanet}). Thus, by combining the loss from each branch ($R$ for RGB, $D$ for depth, and $S$ for shared branches), the overall multi-level loss function $\mathcal{L}_{ml}$ becomes:
\begin{equation}
\mathcal{L}_{ml} = \sum\limits_{i=1}^{5} \lambda_i(\mathcal{L}_i(R) + \mathcal{L}_i(D) + \mathcal{L}_i(S)),
\end{equation}
where $\lambda_i$ is the weight of the different-level loss. To correlate with the network hierarchies, we follow \cite{chen2020global,zhao2020depth} and set the weight $\lambda$ as $\{1, 0.8, 0.6, 0.4, 0.2\}$. 

We expect the multi-level loss to measure the difference between the generated mask and ground truth at various layers, and to force the network to learn hierarchical features that capture long- and short-range spatial relationships between pixels. The gain by adopting the multi-level loss can be found in the ablation study Section \ref{abla:gba} Tab. \ref{tab:abla}.

\section{Experiments}

\subsection{Benchmark Datasets}

To verify the effectiveness of our approach, we firstly train with the conventional training dataset following the protocol presented in \cite{fan2020bbs,zhao2020depth,ji2021calibrated,liu2021TriTrans,zhouiccvspnet} with 2,195 samples: 1,485 samples from the NJU2K-train \cite{ju2014depth} and 700 samples from the NLPR-train \cite{peng2014rgbd}.  For testing, experiments are conducted on five classical benchmark RGB-D datasets. DES \cite{cheng2014depth} : includes 135 images of indoor scenes captured by a Kinect camera. NLPR-test \cite{peng2014rgbd}: contains 300 natural images captured by a Kinect under different illumination conditions. NJU2K-test \cite{ju2014depth}: contains 500 stereo image pairs from different sources such as the Internet, 3D movies, and photographs taken by a Fuji W3 stereo camera, where several depth maps are estimated through an optical flow method \cite{sun2010secrets}. STERE \cite{niu2012leveraging}: includes 1,000 stereoscopic images downloaded from the Internet where the depth map is estimated using the SIFT flow method \cite{liu2010sift}. SIP \cite{fan2019rethinkingd3}: contains 929 images with humans in the scene, and images are acquired by a mobile device. We further evaluate our model on a newly published dataset COME15K \cite{cascaded_cmi} where the depth is estimated through a modified optical flow algorithm \cite{WangZDZJL20}. In this case, our model is trained with provided 8,025 training samples and tested on the ``Difficult'' set with 3,000 images. 

\begin{table*}[t]
\scriptsize
\setlength\tabcolsep{4.3pt}
\setlength\extrarowheight{1pt}
\begin{center}
\caption{\textbf{Quantitative comparison} with SOTA models. $\uparrow$ ($\downarrow$) denotes that the higher (lower) is better. We use the Mean Absolute Error ($M$), max F-measure ($F_m$), S-measure ($S_m$), and max E-measure ($E_m$) as evaluation metrics. (\textbf{Bold}: best.)}
\label{tab:quant}
\begin{tabular*}{\linewidth}{ll|| llll|llll|llll|llll|llll}
 \hline 

\hline

\hline
 Dataset &Size  &\multicolumn{4}{c}{DES} & \multicolumn{4}{c}{NLPR}  & \multicolumn{4}{c}{NJU2K}  & \multicolumn{4}{c}{STERE} & \multicolumn{4}{c}{SIP}  \\
\cline{3-6} \cline{7-10} \cline{11-14} \cline{15-18} \cline{19-22} 

Metric & Mb & $M\downarrow$ &  $F_{\beta}\uparrow $  &  $S_m\uparrow$ &  $E_m\uparrow$  &  $M\downarrow$ &  $F_{\beta}\uparrow $  &  $S_m\uparrow$ &  $E_m\uparrow$  &  $M\downarrow$ &  $F_{\beta}\uparrow $  &  $S_m\uparrow$ &  $E_m\uparrow$  &  $M\downarrow$ &  $F_{\beta}\uparrow $  &  $S_m\uparrow$ &  $E_m\uparrow$  &  $M\downarrow$ &  $F_{\beta}\uparrow $  &  $S_m\uparrow$ &  $E_m\uparrow$  \\
\hline

\multicolumn{15}{l}{\textbf{Performance of RGB-D Models with VGG Backbones}} \\

$DMRA_{19}$ \cite{piao2019dmra} & 278& .030 & .907 & .900 &.934 & .031 & .888 & .899 &.940 
                                & .051 & .896 & .886 &.920 
                                & .047 & .895 & .886 &.930
                                & .086 & .852 & .806 &.847  \\
$A2dele_{20}$ \cite{piao2020a2dele} & 116 & .029 & .897 & .886  &.917 & .029 & .895 & .899  &.943 & .051 & .890 & .871  &.914 & .044 & .892 & .879  &.926 & .070 & .856 & .829  &.887 \\

$ATSA_{20}$ \cite{zhang2020asymmetric}& 131  & .022 & .931 & .917  &.954 & .027 & .907 & .909  &.947 & .046 & .905 & .885  &.928 & .038 & .912 & .896  &\textbf{.940} & .063 & .884 & .849  &.895 \\

$CMMS_{20}$ \cite{li2020rgb}& 546  & .018  & .934 & .934  &.958 & .028 & .914 & .919  &.946 & .044 & .905 & .900  &.929 & .045 & .899 & .894  &.925 & .058 & .893 & .872  &.901 \\
$DANet_{20}$ \cite{zhao2020single}& 128  & .029 & .916 & .904 &.932 & .047 & .904 & .897 &.926 
                                & .045 & .910 & .899 & .927
                                & .048 & .895 & .892 &.919 
                                & .054 & .900 & .888 &.912 \\

$CMWNet_{20}$ \cite{li2020cross}  & 327 & .022 & .939 & .934  &.959 & .029 & .913 & .917  &.941 & .046 & .913 & .903  &.925 & .043 & .911 & .905  &.930 & .062 & .889 & .867  &.901 \\

$HDFNet_{20}$ \cite{pang2020hierarchical}& 308 & .021 & .932 & .926  &.962 & .023 & .926 & .923  &.957 & .039 & .922 & .908  &.939 & .042 & .910 & .900  &.933 &.048 & \textbf{.909} & .886  &.924 \\

$PGAR_{20}$ \cite{chen2020progressively}& 62 & .032 & .894 & .886  &.906 & .027 & .912 & .917  &.941 & .042 & .918 & .909  &.932 & .045 & .902 & .894  &.919 &.072 & .852 & .838  &.875 \\

$SSF_{20}$ \cite{zhang2020select} & 126 & .026 & .912 & .904  &.930 & .027 & .912 & .915  &.947 & .043 & .911 & .899  &.929 & .065 & .859 & .837  &.882 &.091 & .810 & .799  &.855 \\

$CASGNN_{20}$ \cite{luo2020cascade}& 160  & .027 & .917 & .893  &.926 & .025 & .914 & .919  &.953 & .036 & .927 & .910  &.944 & \textbf{.038} & .913 & .899  &\textbf{.940} & -& - & -  &- \\

$D3Net_{21}$ \cite{fan2019rethinkingd3}& 518 & .031 & .909 & .897 &.923 & .030 & .907 & .912 &.942 
                                & .049 & .910 & .900 &.928 
                                & .039 & .911 & .902 &\textbf{.940} 
                                & .063 & .886 & .866 &.897\\
$CDINet_{21}$ \cite{CDINet} & 217 & .020 & .943 & \bl{.937} &.962 & .024 & .923 & .927 &.953 
                                & \textbf{.030} & .928 & \textbf{.918} &.945 
                                & .040 & .912 & \textbf{.913} &.937 
                                & .054 & .904 & .875 &.908 \\

$UCNet_{21}$ \cite{zhang2021uncertainty} & 120 & .018 & .936 & .934 & \textbf{.970} &.025 & .915 & .920 & .953 & .043 & .908 & .897 & .932 & .039 & .908 & .902 &.938 & .051 & .896 & .875 & .915\\ 
$DRLF_{21}$ \cite{wang2020data} & 351 & .030 & .909 & .895  &.918 & .032 & .904 & .903  &.929 & .055 & .896 & .886  &.914 & .050 & .897 & .888  &.916 & .071 & .869 & .850  &.882 \\
$HAINet_{21}$ \cite{li2021hierarchical}  & 228 & .018 & \textbf{.945} & .935  &.967 &  .024&   .920&   .924&   .956 
                                      &  .037&   .924&   .911&   .940
                                      &  .040&   \textbf{.917}&   .907&   .938 
                                      &  .052&   .907&   .879&   .917 \\

$BIANet_{21}$ \cite{zhang2021bilateral} & 189 & .020 &.939 &.931 &.955 & .025 & .921 & .925  &.954 & .039 & .928 & .915  &.939 & .043 & .910 & .903  &.932 & .052 & .904 & .883  &.916\\
                                      
$DCMF_{22}$ \cite{wang2022learning} & 78& .022 & .934 & .932 & .956  &.029 & .913 & .922 & .940 & .041  & .911 &  .902 & .935 & .043 & .916 & .910 & .928 & - & -& -& -\\ 

\rowcolor[RGB]{235,235,250} 
\textbf{Ours (VGG16)} & 269 & \textbf{.017} & .944 & .929 & .968  &\textbf{.021} & \textbf{.927} & \textbf{.928} & \textbf{.962} & .034 & \textbf{.930 }& \textbf{.918} & \textbf{.947} & .039 & .915 & .902 & .939 & \textbf{.045} & \textbf{.909} & \textbf{.889} & \textbf{.927} \\

\hline
\multicolumn{15}{l}{\textbf{Performance of RGB-D Models with ResNet Backbones}} \\
$JLDCF_{21}$ \cite{fu2021siamese} & 548 & .020 & .934 & .931  &.961 & .022 & .925 & .925  &.955 & .041 & .912 & .902  &.936 & .040 & .913 & .903  &.934 & .049 & .903 & .880  &.918 \\

$RD3D_{21}$ \cite{chen2021rgb}  & 179 & .019 & .941 & .935  &.965 & .022 & .927 & .930  &.959 & .036 & .923 & .916  &.941 & .037 &.917 & .911  &.939 & .048 & .906 & .885  &.918\\

$BIANet_{21}$ \cite{zhang2021bilateral} & 244 & .020 & .939 & .930  &.958 & .023 & .924 & .926  &.956 & .036 & .929 & .917  &.942 & .039 & .912 & .905  &.935 & .047 & .904 & .887  &.920 \\

$CoNet_{20}$ \cite{ji2020accurate}& 162 & .024 & .920 & .914  &.944 & .027 & .903 & .911  &.943 & .046 & .902 & .896  &.926 & .037 & .909 & .905  &.941 & .058 & .887 & .860  &.911\\

$DASNet_{20}$ \cite{zhao2020depth} & 141 & .024 & .926 & .905  &.932 & .021 & \bl{.929} & .929  &.960 & .042 & .911 & .902  &.935 & .037 & .915 & .910  &.939 &.051 & .900 & .877  &.918 \\

$BBSNet_{21}$ \cite{zhai2021bifurcated}& 200 & .021 & .942 & .934  &.955 & .023 & .927 & .930  &.953 & .035 & .931 & .920  &.941 & .041 & .919 & .908  &.931 &.055 & .902 & .879  &.910 \\

$DCF_{21}$ \cite{ji2021calibrated} & 435 & .024 &.910 & .905 &.941   &  .022&   .918&   .924&   .958
                                      &  .036&   .922&   .912&   .946
                                      &  .039&   .911&   .902&   .940
                                      &  .052&   .899&   .876&   .916 \\

$DSA2F_{21}$ \cite{sun2021deep}& - & .021 & .896 & .920  &.962 & .024 & .897 & .918  &.950 & .039 & .901 & .903  &.923 & .036 & .898 & .904  &.933 & - & - & -  &- \\



$DSNet_{21}$ \cite{wen2021dynamic} & 661& .021 & .939  & .928 & .956 & .024 & .925 & .926 & .951 & .034 & .929 & .921 &.946 & .036 & .922 & \bl{.914} & .941 & .052 & .899 & .876 & .910\\

$UTANet_{21}$ \cite{zhao2021rgb} & 186 & .026 & .921 & .900  &.932 & \bl{.020} & .928 & \bl{.932}  &\bl{.964} & .037 & .915 & .902  &.945 & .033 & .921 & .910  &.948 &.048 & .897 & .873  &.925 \\

$C2DFNet_{22}$ \cite{zhang2022c} & 198& .020 & .937 & .922 & .948 &.021 & .926 & .928 & .956 & - & -& -& -   & .038 & .911 & .902 & .938 & .053 & .894 & .782 & .911\\

$MVSalNet_{22}$ \cite{zhou2022mvsalnet}   & - & .019 & .942& .937& \bl{.973}  &  .022&   .931&   .930&   .960 &  .036&   .923&   .912&   .944  &  .036&   .921&   .913&   .944 & - & - &- &-\\  

$SPSN_{22}$ \cite{lee2022spsn}  & 149 & .017 & .942& .937& \bl{.973} &  .023&   .917&   .923&   .956  &  .032&   .927&   .918&   .949 &  .035&   .909&   .906&   .941  &  .043&   .910&   .891&   \bl{.932} \\

\rowcolor[RGB]{235,235,250} 


\textbf{Ours (ResNet50)} & 523 & \bl{.015} & \bl{.947} & \bl{.939} & \bl{.973} & .022 & .927 & .925 & .957 & \bl{.030} & \bl{.937} & \bl{.924} & \bl{.952} & \bl{.033} & \bl{.926} & \bl{.914} & \bl{.948} & \bl{.043} & \bl{.915} & \bl{.893} & .930 \\

\hline
\multicolumn{15}{l}{\textbf{Performance of RGB-D Models with Res2Net Backbones}}  \\

$BIANet_{21}$ \cite{zhang2021bilateral} & 244 & .017 & .948 &.942 &.972 & .022 & .926 & .928  &.957 & .034 & .932 & .923  &.945 & .038 & .916 & .908  &.935 & .046 & .908 & .889  &.922  \\

$SPNet_{21}$ \cite{zhouiccvspnet}& 702& .014 & .950 & .945  & \bl{.980} &  \bl{.021} &  .925 &  .927  &.959 & \bl{.028} & .935 & .925  &\bl{.954} & .037 & .915 & .907  &.944 & \bl{.043} & .916 & \bl{.894}  &\bl{.930} \\ 


\rowcolor[RGB]{235,235,250} 

\textbf{Ours (Res2Net50)}&  525 & \bl{.013} & \bl{.952} & \bl{.946}  &\bl{.980} & \bl{.021} & \bl{.929} & \bl{.930}  &\bl{.961} &  .029 & \bl{.939} & \bl{.926}  &\bl{.954} & \bl{.035} & \bl{.921} & \bl{.911}  &\bl{.946} & \bl{.043} & \bl{.919} &  .892  & .927\\
 




\hline





 \hline 

\hline

\hline
\end{tabular*}
\end{center}

\end{table*}

\begin{figure}[t]
\centering
\includegraphics[width=.9\linewidth,keepaspectratio]{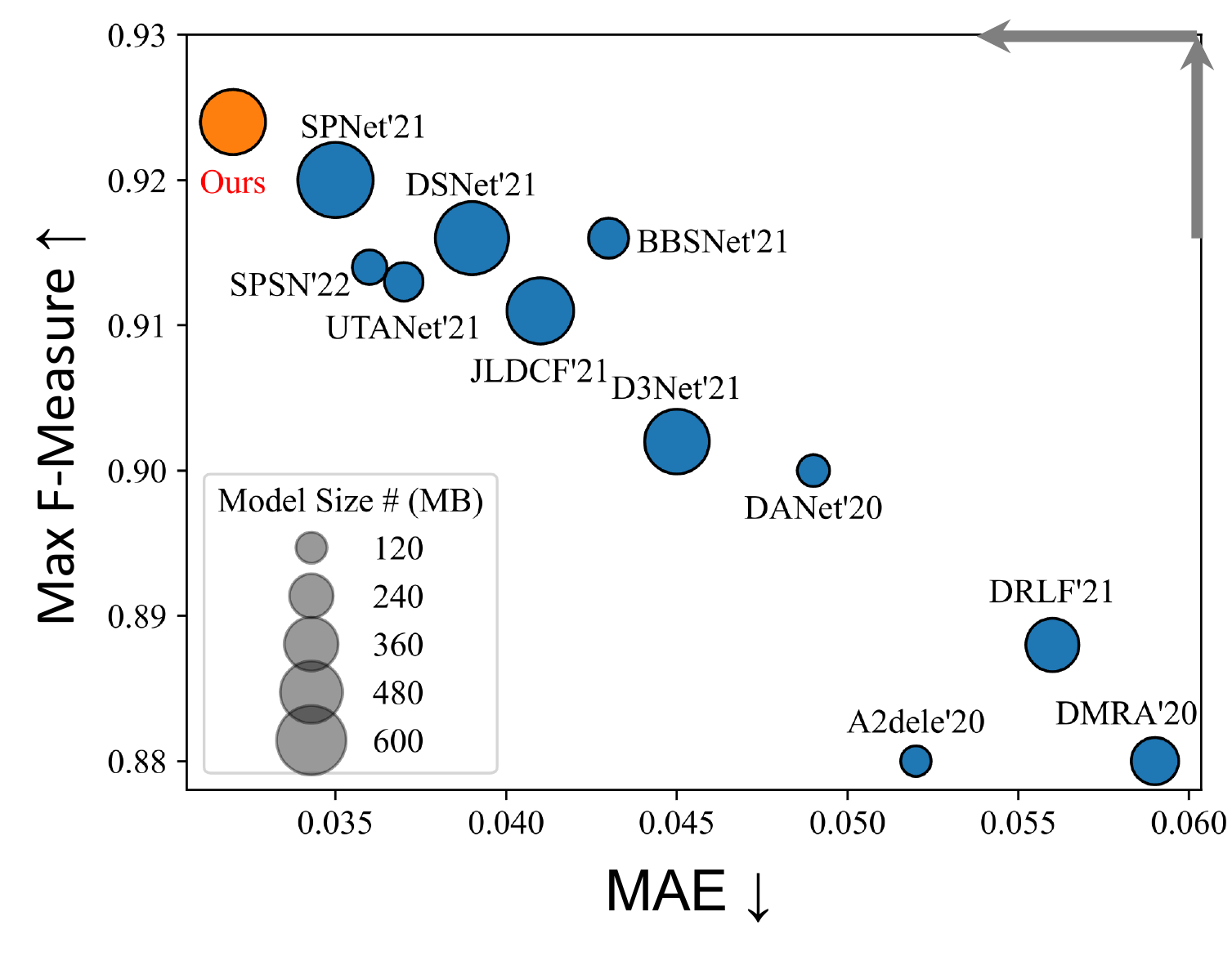}
\caption{Average \textbf{Max F-Measure, MAE, and Model Size} of different methods on benchmark datasets. The circle size denotes the model size. Note that better models are shown in the upper left corner (i.e., with a larger F-measure and smaller MAE). Methods with smaller size perform inferior, making our method both efficient and accurate.}
\label{fig:average}

\end{figure}

\subsection{Experimental Settings}
Our model is implemented based on Pytorch and trained with a V100 GPU. Our backbone is initialized with the pre-trained weights obtained from ImageNet. For the depth stream, we modify the first convolution to start from one channel. The input RGB-D resolution is fixed to 352$\times$352. We choose the Adam algorithm as our optimizer. We initialize the learning rate to be 1$e-$4 which is further divided by 10 every 60 epochs. The total training time takes around 6 hours for 100 epochs.  During training, we adopt random flipping, rotating, and border clipping for data augmentation. During inference, the prediction maps from the shared branch are the final outputs (middle branch of Fig. \ref{fig:hidanet}).

We evaluate our performance with four generally-recognized metrics: F-measure is a region-based similarity metric that takes into account both Precision (P) and Recall (R). Mathematically, we have : $F_{\beta} = \frac{(1+\beta^2) \cdot P \cdot R}{ \beta^2 \cdot P + R}$. The value of $\beta^2$ is set to be $0.3$ as suggested in \cite{achanta2009frequency} to emphasize the precision. In this paper, we report the \textbf{maximum F-measure} ($F_{\beta}$) score across the binary maps of different thresholds. \textbf{Mean Absolute Error ($M$)} measures the approximation degree between the saliency map and ground-truth map at the pixel level.
\textbf{S-measure ($S_m$)}  \cite{Cheng2021sMeasure} evaluates the similarities between object-aware ($S_o$) and region-aware ($S_r$) structures of the saliency map compared to the ground truth. Mathematically, we have: $S_m= \alpha \cdot S_o + (1 - \alpha) \cdot S_r$, where $\alpha$ is set to be $0.5$.
\textbf{E-measure ($E_m$)} evaluates both image-level statistics and local pixel-matching information. Mathematically, we have: $E_m= \frac{1}{W\times H} \sum_{i=1}^{W} \sum_{j=1}^{H} \phi_{FM}(i,j)$, where $\phi_{FM}(i,j)$ stands for the enhanced-alignment matrix as presented in \cite{fan2018enhanced}. To make a fair comparison, we use the same protocol as \cite{zhouiccvspnet} to evaluate the officially released saliency maps for each SOTA method.

\subsection{Comparison with SOTA RGB-D models}
\label{quantitative}

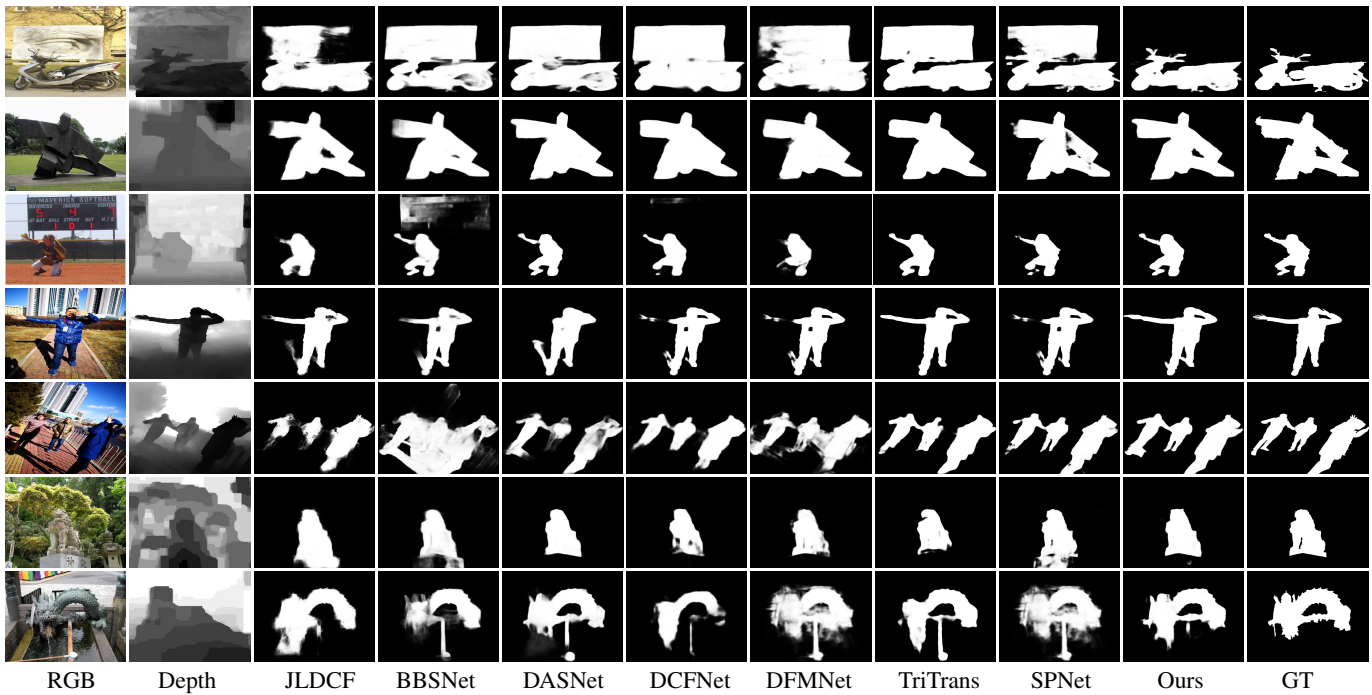
\begin{figure*}[t]
\centering
\includegraphics[width=\linewidth,keepaspectratio]{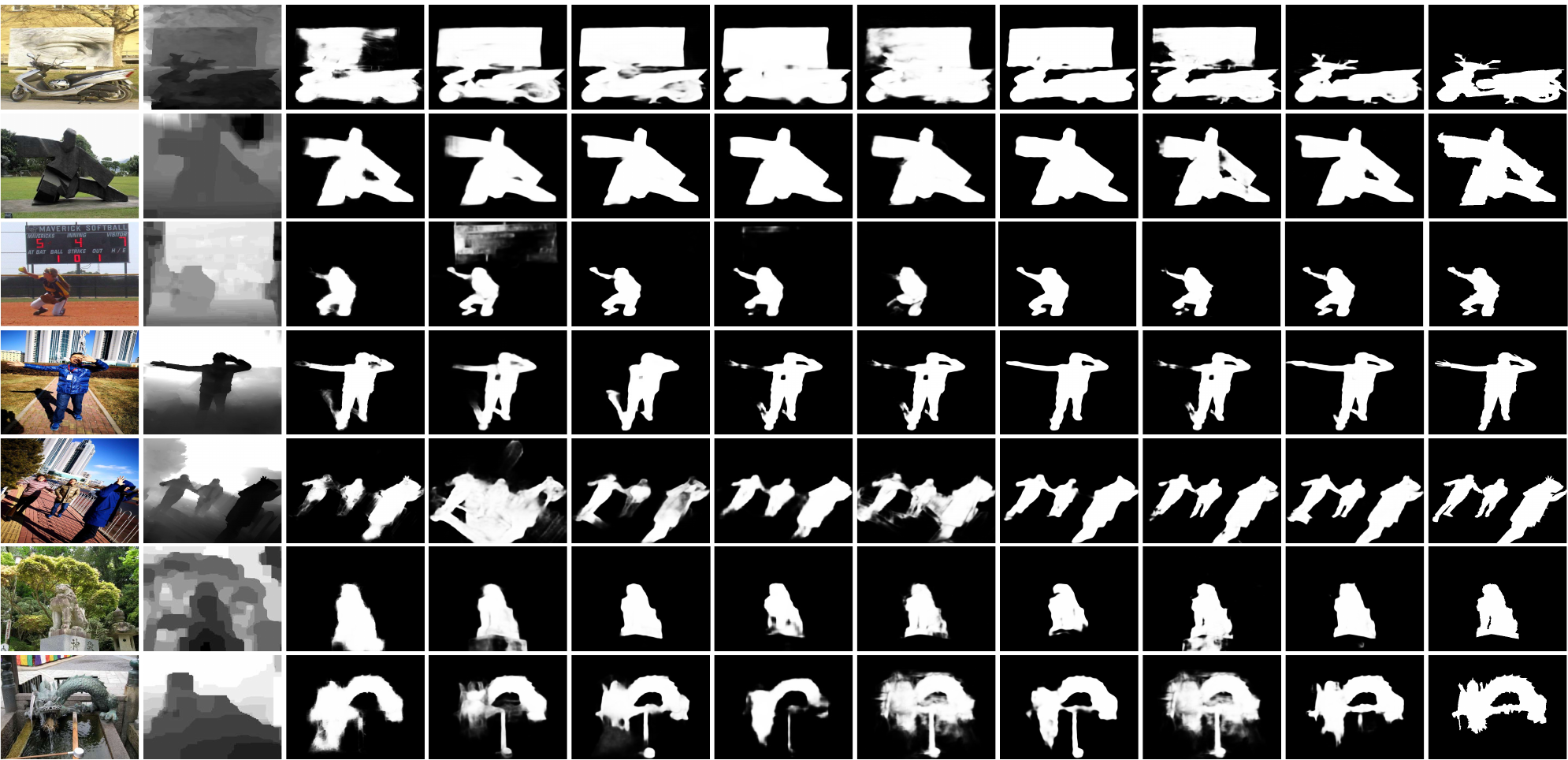}
  \put(-500,-10){{\color{black}{\small{RGB}}}}
  \put(-458,-10){{\color{black}{\small{Depth}}}}
  \put(-410,-10){{\color{black}{\small{JLDCF}}}}
 \put(-368,-10){{\color{black}{\small{BBSNet}}}}
  \put(-320,-10){{\color{black}{\small{DASNet}}}}
  \put(-272,-10){{\color{black}{\small{DCFNet}}}}
  \put(-228,-10){{\color{black}{\small{DFMNet}}}}
  \put(-178,-10){{\color{black}{\small{TriTrans}}}}
  \put(-128,-10){{\color{black}{\small{SPNet}}}}
  \put(-80,-10){{\color{black}{\small{Ours}}}}
  \put(-33,-10){{\color{black}{\small{GT}}}}
  
\caption{\textbf{Visual comparison} between our HiDAnet and SOTA methods in various challenging cases. It can be seen that our method better explores the granularity prior to reason about the saliency map closer to the ground truth.}
\label{fig:quali}
\end{figure*}

\noindent\textbf{Quantitative Comparison:}  We provide in Figure \ref{fig:average} an overview of the average performance on conventional benchmark datasets, i.e., DES \cite{cheng2014depth}, NLPR \cite{peng2014rgbd}, NJU2K \cite{ju2014depth}, STERE \cite{niu2012leveraging}, and SIP \cite{fan2019rethinkingd3}. The detailed quantitative performances can be found in Tab. \ref{tab:quant}. We also present in Tab. \ref{tab:come} the quantitative comparison on the newly published challenging COME15K \cite{cascaded_cmi} dataset. All saliency maps are directly provided by authors or computed by authorized codes.

\begin{table*}[t]
\scriptsize
\setlength\tabcolsep{0.5pt}
\setlength\extrarowheight{1pt}
\begin{center}
\caption{\textbf{Quantitative comparison} on the challenging COME15K \emph{Difficult} test set~\cite{cascaded_cmi}. We use the Mean Absolute Error ($M$), max F-measure ($F_m$), S-measure ($S_m$), and max E-measure ($E_m$) as evaluation metrics. (\textbf{Bold}: best.)}
\label{tab:come}
\begin{tabular*}{0.9\textwidth}{@{\extracolsep{\fill}}*{9}{c}}
 \hline 

\hline

\hline
& $JLDCF$ & $A2dele$   & $DMRA$  & $CoNet$ & $BBSnet$ & $SPnet$ &$CMINet$ & $\textbf{Ours}$ \\
\hline
$M \downarrow$    
& .075 & .092 & .137 & .113 & .071 &.065 &.064 & \bl{.062} \\
$E_m \uparrow$      
&.870 &.838  &.775 &.813  & .876 &.888 & \bl{.893} & \bl{.893}\\
 \hline 

\hline

\hline
\end{tabular*}
\end{center}
\end{table*}

\begin{figure*}[t]
\centering
\includegraphics[width=0.9\linewidth,keepaspectratio]{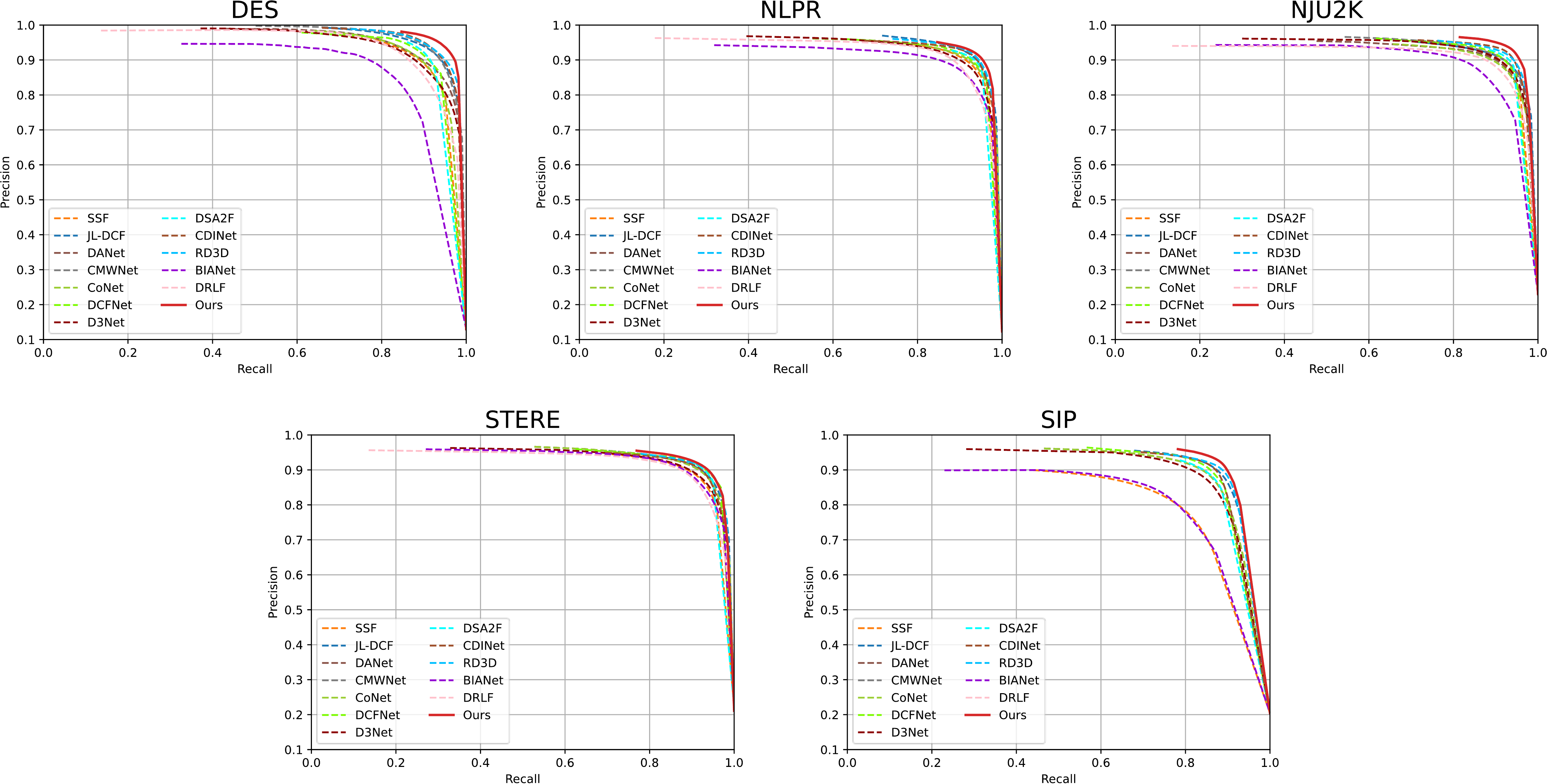}
\caption{Comparison on \textbf{PR curves}. Our HiDANet achieves better performance compared to the 12 listed SOTA methods across different datasets.}
\label{fig:curves}
\end{figure*}

Under the consideration of a fair comparison, we conduct experiments with different backbones such as VGG16 \cite{simonyan2014vgg}, ResNet50 \cite{He2016Residual}, and Res2Net50 \cite{gao2019res2net}. It can be seen that our HiDAnet with each backbone achieves comparable and superior performance compared to the SOTA models with the same backbone. Specifically, our HiDANet with VGG16 backbones achieves significantly better performance on NLPR and SIP datasets, while being very competitive on the model size with 269 MB and around 6 FPS. Our HiDAnet with ResNet50 backbones further sets new SOTA records on DES, NLPR, and NJU2K datasets with 523 MB and around 12 FPS. We also follow the SOTA SPNet and replace our backbone with Res2Net50. It can be seen that our method performs favorably compared to SPNet with only 525 MB compared to that of SPNet with 702 MB. Our FPS is around 11. We also exhibit in Fig. \ref{fig:curves} 
the PR curves with several latest published models to further demonstrate the superior performance of our model.

Finally, in addition to the difference in the backbone, we observe that existing works adopt different architectures, i.e.,  design of decoder, supervision, training settings, etc. Under the consideration of fair comparison and to purely analyze the effectiveness of encoder fusion design, we re-implement several fusion alternatives under the same architecture (Res2Net50 + fusion). Specifically, we choose the same backbone (Res2Net50), the same decoder (the SOTA \cite{zhouiccvspnet}), loss (multi-scale supervision), and the same training settings as ours. The only difference between one model to another is in the fusion module. The quantitative comparison can be found in Table \ref{tab:fusion}. It can be seen that by replacing our fusion with other methods, the empirical results significantly drop. This validates the superior effectiveness of our granularity and CDA in leveraging RGB-D cues compared to other alternatives.

\begin{table}[t]
\scriptsize
\setlength\tabcolsep{2.8pt}
\setlength\extrarowheight{1pt}\begin{center}
\caption{Quantitative comparison with different \textbf{fusion designs}. We replace our fusion module with four SOTA fusion modules and retrain the new networks under the same training setting. We use the Mean Absolute Error ($M$), max F-measure ($F_m$), S-measure ($S_m$), and max E-measure ($E_m$) as evaluation metrics. (\textbf{Bold}: best.)}
\label{tab:fusion}
\begin{tabular}[t]{lc || c c| c c| c c| c c }
\hline 

\hline

\hline
 Dataset &Size  & \multicolumn{2}{c|}{NLPR}  & \multicolumn{2}{c|}{NJU2K}  & \multicolumn{2}{c|}{STERE} & \multicolumn{2}{c}{SIP}  \\
\cline{3-4} \cline{5-6} \cline{7-8} \cline{9-10}  

Metric & Mb & $F_{\beta}\uparrow $  &    $E_m\uparrow$  &  $F_{\beta}\uparrow $  &   $E_m\uparrow$  &   $F_{\beta}\uparrow $  &    $E_m\uparrow$  &   $F_{\beta}\uparrow $  &   $E_m\uparrow$  \\
\hline

\hline
\rowcolor[RGB]{235,235,250} 

 Res2Net50 + Ours &525  & \bl{.929}  &\bl{.961} & \bl{.939}  &\bl{.954} & {.921}  &\bl{.946} & \bl{.919}   &{.927} \\ 
 \hline
Res2Net50 + BBS \cite{fan2020bbs} & 509 & .922 &.953 &.918 &.939 &.890 &.909 &.916 &.917 \\ 
Res2Net50 + CDI \cite{CDINet} & 531 & .926 &.958 &.927 &.946 &\bl{.922} &.945 &.907 &.920 \\
Res2Net50 + DCF \cite{ji2021calibrated} & 347& .927 &.958 &.933 &.948 &.916 &.939 &.911 &.923 \\
Res2Net50 + SP \cite{zhouiccvspnet} & 737& .925 &.959 &.935 &\bl{.954} &.915 &.944 &.916 & \textbf{.930}  \\
 \hline 

\hline

\hline
\end{tabular}
\end{center}
\end{table}

\noindent\textbf{Qualitative Comparison: } Fig. \ref{fig:quali} illustrates generated saliency maps of different methods on challenging cases: cluttered background and foreground with a similar appearance ($1^{st}-2^{nd}$ rows), human in the scene ($3^{rd} - 5^{th}$ rows), and low contrast on the depth map  ($6^{th} - 7^{th}$ rows). Compared to the SOTA models, our HiDAnet yields results closer to the ground-truth masks. For the motorbike in the $1^{st}$ row, our model can selectively remove the background region (board). For the sculpture in the $2^{nd}$ row, our network pays local attention to the foreground and thus the hollow part can be detailed. We can also accurately extract the human with large deformations ($3^{rd} - 5^{th}$ rows). 

\label{rob}

\begin{table*}[t]
\scriptsize
\setlength\tabcolsep{0.5pt}
\setlength\extrarowheight{1pt}
\begin{center}
\caption{Experiments under \textbf{inferior conditions} with simulated depth noises ($RMSE$, $\delta 1$). While $RMSE$, $\delta 1$ are 0, it represents the result without simulated noises. \textbf{Drop} $\Delta$ denotes the absolute performance difference. Our HiDAnet leads to a more stable performance compared to the SOTA methods with a lower $\Delta$ under different inferior conditions, proving that our model is more robust against depth noises. We use the Mean Absolute Error ($M$), max F-measure ($F_m$), S-measure ($S_m$), and max E-measure ($E_m$) as evaluation metrics. (\textbf{Bold}: best.)}
\label{tab:rob}
\begin{tabular*}{\textwidth}{@{\extracolsep{\fill}}*{21}{c}}
 \hline 

\hline

\hline
Dataset &\multicolumn{6}{c}{DES} & \multicolumn{6}{c}{NLPR}  & \multicolumn{6}{c}{NJU2K} \\

\cline{2-7} \cline{8-13} \cline{14-19}

Metric & $RMSE$ & $\delta 1$  & $M\downarrow$ &  $F_{\beta}\uparrow $  &  $S_m\uparrow$ &  $E_m\uparrow$ & $RMSE$ & $\delta 1$  &  $M\downarrow$ &  $F_{\beta}\uparrow $  &  $S_m\uparrow$ &  $E_m\uparrow$  & $RMSE$ & $\delta 1$ &  $M\downarrow$ &  $F_{\beta}\uparrow $  &  $S_m\uparrow$ &  $E_m\uparrow$   \\
 \hline 
 
 
 $CMINet_{21}$ & 0 & 0 & .016 & .944 & .940  &.975  & 0 & 0  & .020 & .931& .932  &.959  & 0 & 0 & .028 & .940& .929 & .954 \\ 
 
$CMINet_{21}$ & .261 & .270  & .022 & .925 & .920 & .952 

& .259 & .342 & .021 &.929 & .932 & .960 

& .236 & .413 &.032 &.934 &.922 &.948 \\

Drop $\Delta(\%)$  & - & - &.6 & 1.9 &2.0   &2.3   &- & - &.1 & 0.2 & \textbf{0}  &\textbf{.1}  & - & - & 0.4 & 0.6 & .7  &.6\\ 

 \hline 
$SPNet_{21}$ & 0 & 0 &.014 & .950 & .945  &.980   &0 & 0 &.021 & .925 & .927  &.959  & 0 & 0 & .028 & .935 & .925  &.954\\ 

$SPNet_{21}$ & .261 & .270  & .017 & .944 & .935 & .972

& .259 & .342 & .020 &.922 & .924 & .956 

& .236 & .413 &.033 &.931 &.920 &.946 \\

Drop $\Delta(\%)$  & - & - &.3 & .6 &1   &.8   &- & - &.1 & .3 & .3  &.3  & - & - & .5 & \bf{.4} & .5  &.8\\

\hline
$Ours$ & 0 & 0& .013 & .952 &.946  &.980   & 0 & 0  & .021 & .929 & .930  &.961  & 0 & 0 & .029 & .939 & .926  &.954 \\ 

$Ours$  & .261 & .270  & .015 & .948 & .943 & .980

& .259 & .342 & .021 &.930 & .930 & .962

& .236 & .413 &.029 & .935 &.925 &.953 \\

Drop $\Delta(\%)$  & - & - &\bf{.2} & \bf{.4} &\bf{.3}   &\bf{0}   &- & - &\bf{0} & \bf{.1} & \bf{0}  &\bf{.1}  & - & - & \bf{0} & \bf{.4} & \bf{.1}  &\bf{.1}\\ 

\hline

\hline
\end{tabular*}
\end{center}
\end{table*}

\noindent\textbf{Robustness against Depth Noise:} Tab. \ref{tab:rob} reports the robustness analysis on the depth quality. To make a fair comparison, we conduct experiments and compare with the SOTA SPnet \cite{zhouiccvspnet} and CMINet \cite{cascaded_cmi} under the same inferior condition with a simulated Gaussian noise on depth. We further evaluate the performances on the simulated noisy testing dataset. The noise level is defined by the conventional metrics $RMSE$ and $\delta1$. While $RMSE$ and $\delta1$ are 0, we report the performance tested with the vanilla dataset (without noise). \textbf{Drop} $\Delta$ denotes the performance degradation by \% under the simulated depth noise. 


Note that CMINet designs a multi-scale mutual information minimization during the encoding stage and lately merge multi-modal features at the semantic level, yielding an unsatisfactory performance while dealing with noisy datasets (drop 2.0\% $S_m$  and 2.3\% $E_m$ for noisy DES). Differently, both SPnet and ours fuse features at each stage, leading to superior robustness against the noise. Compared to SPnet, it can be seen that our performance is more stable, which can be attributed to our granularity attention and fusion designs. The gain of each component can be found in Tab. \ref{tab:abla}.

\section{Ablation Study}

\begin{table*}[t]
\scriptsize
\begin{center}
\caption{Ablation study on attention designs with \textbf{different average pooling} methods. We use the Mean Absolute Error ($M$), max F-measure ($F_m$), S-measure ($S_m$), and max E-measure ($E_m$) as evaluation metrics. (\textbf{Bold}: best.)}
\label{tab:reb}
\begin{tabular*}{\textwidth}{@{\extracolsep{\fill}}*{10}{c}}
 \hline 

\hline

\hline
\multirow{2}{*}{$\#$} & \multirow{2}{*}{$Description$} &\multicolumn{2}{c}{DES}  & \multicolumn{2}{c}{NLPR}  & \multicolumn{2}{c}{NJU2K}  & \multicolumn{2}{c}{STERE} \\
\cline{3-4} \cline{5-6} \cline{7-8} \cline{9-10}
& &$M\downarrow$ &  $F_{\beta}\uparrow $ &$M\downarrow$ &  $F_{\beta}\uparrow $  &$M\downarrow$ &  $F_{\beta}\uparrow $ &$M\downarrow$ &  $F_{\beta}\uparrow $
\\
\hline
I & Vanilla Global Attention + Global Pooling& .019 & .940 & \bf{.020} & \bl{.929} &.030 & .936 & .037 & .918 \\
II &Local Attention + Global Pooling & .015 & .947 & .021 & .927 &.032 & .928 & .038 & .915 \\
\bl{III} & Our Local Attention + Local Pooling & \bl{.013} & \bl{.952} &  .021 & \bl{.929} &\bl{.029} & \bl{.939} & \bl{.035} & \bl{.921} \\
 \hline 

\hline

\hline
\end{tabular*}
\end{center}
\end{table*}

\label{abla:gba}
\subsection{Comparison with Vanilla Channel Attention} We propose granularity-based attention (GBA) referring to geometric priors, which differs from the traditional channel attention on the pooling strategies. Formally, let $z\in \mathbb{R}^C$ be the squeezed spatial information from feature $x\in \mathbb{R}^{H\times W\times C}$. Accordingly, we can obtain three variations of average pooling: 
\begin{equation}
\begin{split}
   &(I)\ z = \frac{\sum\sum x(.)}{H\times W}; \\
   &(II)\  z = \frac{\sum\sum x(.) \cdot m_i()}{H\times W}; \\
   &(III) \ z = \frac{\sum\sum x(.) \cdot m_i(.)}{\sum\sum m_i(.)} \\
\end{split}
\end{equation}
where (I) denotes the vanilla global average pooling, (II) is the global pooling with local region $m_i(.)$, and (III) is our proposed GBA module that applies local pooling with local region $m_i(.)$. Note that when depth data is constant, i.e., all the pixels belong to the same granularity, our local average becomes the global average pooling and our model is equivalent to the conventional channel attention \cite{hu2018senet,wang2020eca}. To verify our effectiveness, we conduct experiments by replacing our local pooling with the aforementioned poolings.  Empirical results in Tab. \ref{tab:reb} show that compared to (I), (II) can better leverage local awareness which spatially constrains attention around the local region. However, with a large $H\times W$, the attention activation is limited. Hence, we further propose to adopt local pooling to automatically adjust the weight (III). Our superior performance validates the effectiveness of our local design.  

\subsection{Why GBA in both streams} We analyze in Tab. \ref{tab:gbargbd} the contribution of GBA for both RGB and Depth feature modelings: (A) We remove the GBA from our network, denoted as RGB+D; (B) GBA is only applied in the RGB stream, denoted as RGB(G) + D; (C) GBA is applied in both streams, denoted as RGB(G) + D(G). We observe that the performance augments by gradually inserting GBA into the encoders. This shows that GBA can be considered as depth-aware attention for the RGB stream and as a self-enhancement module for the Depth stream to produce regions with favorable objectness. 


\begin{table*}[t]
\scriptsize
\setlength\tabcolsep{4.7pt}
\setlength\extrarowheight{1pt}
\begin{center}
\caption{Experiments by gradually adding \textbf{granularity attention} module on RGB and Depth streams. RGB(G)/D(G) denotes the case when granularity attention is applied to RGB/Depth branch. We use the Mean Absolute Error ($M$), max F-measure ($F_m$), S-measure ($S_m$), and max E-measure ($E_m$) as evaluation metrics. (\textbf{Bold}: best.)}
\label{tab:gbargbd}
\begin{tabular*}{\linewidth}{l|| llll|llll|llll|llll|llll}
 \hline 

\hline

\hline
Dataset &\multicolumn{4}{c}{DES} & \multicolumn{4}{c}{NLPR}  & \multicolumn{4}{c}{NJU2K}  & \multicolumn{4}{c}{STERE} & \multicolumn{4}{c}{SIP}  \\
\cline{2-5} \cline{6-9} \cline{10-13} \cline{14-17} \cline{18-21} 

Metric &  $M\downarrow$ &  $F_{\beta}\uparrow $  &  $S_m\uparrow$ &  $E_m\uparrow$  &  $M\downarrow$ &  $F_{\beta}\uparrow $  &  $S_m\uparrow$ &  $E_m\uparrow$  &  $M\downarrow$ &  $F_{\beta}\uparrow $  &  $S_m\uparrow$ &  $E_m\uparrow$  &  $M\downarrow$ &  $F_{\beta}\uparrow $  &  $S_m\uparrow$ &  $E_m\uparrow$  &  $M\downarrow$ &  $F_{\beta}\uparrow $  &  $S_m\uparrow$ &  $E_m\uparrow$  \\
\hline
(A) RGB + D & .015 & .949 & .940  &.972 & .022 & .925 & .927  &.960 & .030 & .932 & .923  &.952 & .037 & .913 & .901  &.936 & .046 & .914 & .889  &.923 \\
(B) RGB(G) + D& .014 & .951 & .943  &\bl{.980} & \bl{.021} & .927 & .926  &.960 & .030 & .936 & .923  &.953 & .036 & .916 & .907  &.945 & \bl{.043} & \bl{.919} & \bl{.894}  &\bl{.928} \\

\rowcolor[RGB]{235,235,250} 

(C) RGB(G) + D(G)& \bl{.013} & \bl{.952} & \bl{.946}  &\bl{.980} & \bl{.021} & \bl{.929} & \bl{.930}  &\bl{.961} & \bl{.029} & \bl{.939} & \bl{.926}  &\bl{.954} & \bl{.035} & \bl{.921} & \bl{.911}  &\bl{.946} & \bl{.043} & \bl{.919} & .892  &.927 \\ 
 \hline 

\hline

\hline
\end{tabular*}
\end{center}
\end{table*}

\begin{figure}[ht]
\centering
\includegraphics[width=\linewidth,keepaspectratio]{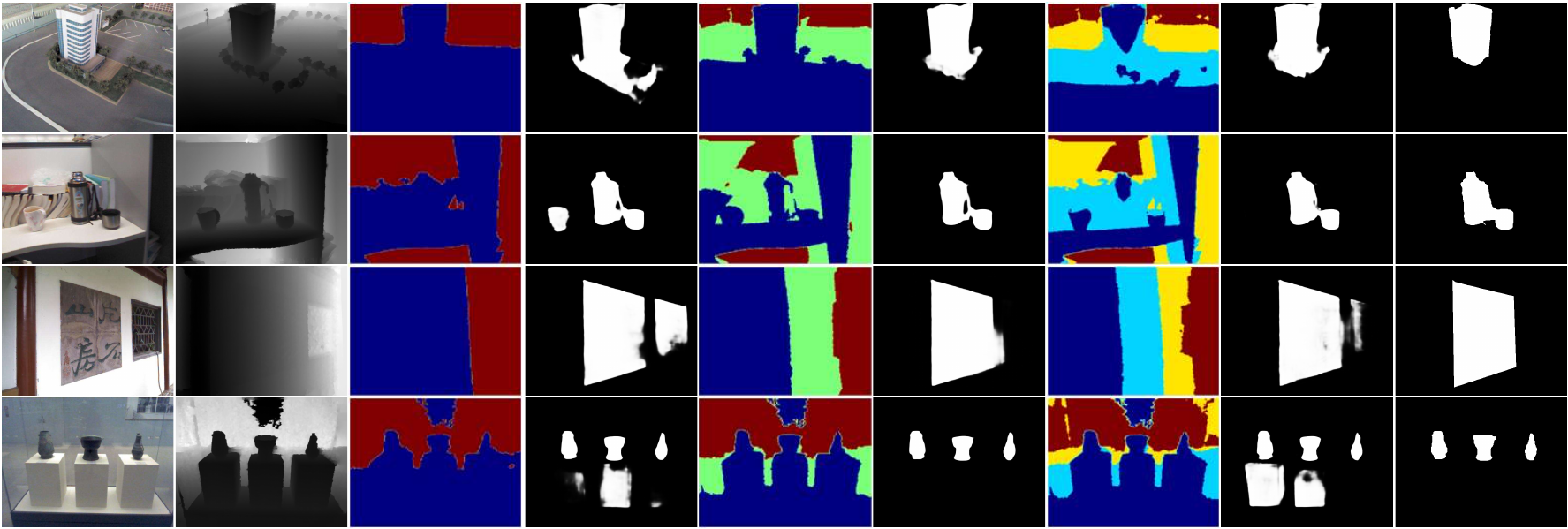}
  \put(-245,-10){{\color{black}{\small{RGB}}}}
  \put(-220,-10){{\color{black}{\small{Depth}}}}
  \put(-193,-10){{\color{black}{\small{$T=1$}}}}
  \put(-165,-10){{\color{black}{\small{$Mask_{1}$}}}}
  \put(-138,-10){{\color{black}{\small{$T=2$}}}}
  \put(-108,-10){{\color{black}{\small{$Mask_{2}$}}}}
  \put(-80,-10){{\color{black}{\small{$T=3$}}}}
  \put(-52,-10){{\color{black}{\small{$Mask_{3}$}}}}
  \put(-21,-10){{\color{black}{\small{GT}}}}
\caption{Qualitative comparison with \textbf{different numbers of Otsu} thresholds ($T=1,2,3$) for our granularity-based attention. With the threshold $T$, we divide the depth map into $T+1$ regions with different colors. Each region shares the same granularity of geometric information. With one threshold $T=1$, the local regions are coarse and cannot get the full benefit from the geometric priors. This results in unsatisfactory salient masks ($4^{th}$ column). With two thresholds $T=2$, the depth map is better discretized with more fine-grained details, yielding salient masks closer to the ground truth ($6^{th}$ column). With three thresholds $T=3$, the depth map is over-discretized, resulting in sub-optimal salient masks ($8^{th}$ column). Our plain HiDAnet is built upon $T=2$.}
\label{fig:otsu}
\end{figure}

\subsection{Number of Otsu Regions for GBA} Our fine-grained details are determined by the number of Otsu regions as shown in Figure \ref{fig:otsu}. The two first columns represent the paired RGB-D inputs. On the $3^{rd}$, $5^{th}$, and $7^{th}$ columns we list the Otsu regions with different numbers of multi granularities, respectively. On the $4^{th}$, $6^{th}$, and $8^{th}$ columns we list the generated masks with different numbers of thresholds $T=1,2,3$, respectively.

By comparing the $3^{rd}$ and $5^{th}$ columns, it can be seen that a small number of Otsu threshold $T=1$ cannot get the full benefit from the geometric priors. For example, the building in the $1^{st}$ row cannot be perfectly distinguished from the background; the cups in the $2^{nd}$ row are mixed with the table and a part of the wall. The unsatisfactory thresholding on the depth histogram leads to sub-optimal performance of granularity-based attention that the discriminatory power cannot be fully exploited. While augmenting the number of thresholds to $T=2$, we observe from the $5^{th}$ column that the scene can be better discretized. The fine-grained details contribute to the clearer boundary generation as shown in the $6^{th}$ column. We further augment the number of thresholds to $T=3$ and observe the over-discretization, leading to the misunderstanding on the depth map. Thus, it results in lower quality salient masks as shown in the $8^{th}$ column. 

Thus, we perform the experiments with different numbers of thresholds $T$. Tab. \ref{tab:otsu} shows that the best overall performance is achieved with $T=2$ thresholds, thus with $n=3$ regions. It can be considered as a scene discretization into three parts: close, middle, and far regions. Our plain HiDAnet is with $T=2$ thresholds and achieves the best performance. We also discover that the sensitivity to thresholding varies from one dataset to another, especially the NLPR dataset which is not highly sensitive to the granularity. This is mainly due to the fact that NLPR contains objects residing in the background. In such circumstances, the target object has the mixed depth response as the background, leading to less-noticeable granularity as shown in the last two rows of Figure \ref{fig:otsu}. In more common and popular cases (DES, NJU2K, STERE, and SIP), our fine-grained details achieve significant improvement compared to our baseline with conventional attention as shown in Tab. \ref{tab:otsu}.


\begin{table*}[t]
\scriptsize
\setlength\tabcolsep{4.9pt}
\setlength\extrarowheight{1pt}
\begin{center}
\caption{Quantitative comparison with \textbf{different Otsu thresholds}. Our plain HiDAnet is with $T=2$ thresholds. $T=2$ achieves the best performance with a reasonable FPS. We use the Mean Absolute Error ($M$), max F-measure ($F_m$), S-measure ($S_m$), and max E-measure ($E_m$) as evaluation metrics. (\textbf{Bold}: best.)}
\label{tab:otsu}
\begin{tabular*}{\linewidth}{ll|| llll|llll|llll|llll|llll}
 \hline 

\hline

\hline
Dataset & &\multicolumn{4}{c}{DES} & \multicolumn{4}{c}{NLPR}  & \multicolumn{4}{c}{NJU2K}  & \multicolumn{4}{c}{STERE} & \multicolumn{4}{c}{SIP}  \\
\cline{3-6} \cline{7-10} \cline{11-14} \cline{15-18} \cline{19-22} 

Metric & $FPS\uparrow$  &  $M\downarrow$ &  $F_{\beta}\uparrow $  &  $S_m\uparrow$ &  $E_m\uparrow$  &  $M\downarrow$ &  $F_{\beta}\uparrow $  &  $S_m\uparrow$ &  $E_m\uparrow$  &  $M\downarrow$ &  $F_{\beta}\uparrow $  &  $S_m\uparrow$ &  $E_m\uparrow$  &  $M\downarrow$ &  $F_{\beta}\uparrow $  &  $S_m\uparrow$ &  $E_m\uparrow$  &  $M\downarrow$ &  $F_{\beta}\uparrow $  &  $S_m\uparrow$ &  $E_m\uparrow$  \\
\hline
$T = 0 $ & \bl{13.3} & .019 & .941 & .927  &.955 & .020 & \bl{.929} & \bl{.931}  &\bl{.961} & .031 & .936 & .924  &.952 & .037 & .919 & .908  &.943 & .046 & .915 & .888  &.924 \\
$T = 1 $ & 12.6 & .015  & .951 & \textbf{.948} &.979 & .023 & .927 & .927  &.960 & \bl{.029} & .933 & .924  &.953 & .035 & .918 & .908  &.944 & .044 & .918 & \bl{.894}  &\bl{.927} \\
\rowcolor[RGB]{235,235,250} 

$T=2 $ & 11.3 & \bl{.013} & \bl{.952} & .946  &\bl{.980} & .021 & \bl{.929} & .930  &\bl{.961} & \bl{.029} & \bl{.939} & \bl{.926}  &\bl{.954} & \bl{.035} & \bl{.921} & \bl{.911}  &\bl{.946} & \bl{.043} & \bl{.919} & .892  &\bf{.927} \\ 
$T = 3 $ & 10.5 & .015 & .949 & .942  &.979 & \bl{.020} & .929 & .928  &\bl{.961} & .031 & .929 & .920  &.949 & .036 & .914 & .900  &.940 & .044 & .916 & .891  &.925 \\

 \hline 

\hline

\hline
\end{tabular*}
\end{center}
\end{table*}

\begin{table}[t]
\scriptsize
\setlength\tabcolsep{2.2pt}
\setlength\extrarowheight{1pt}
\begin{center}
\caption{Ablation study on \textbf{key components} of our proposed HiDAnet. We partially remove key components or replace the fusion designs with a simple addition. $Skip$ stands for the skip connection with the proposed cross dual attention. $\mathcal{L}_{ml}$ denotes the multi-level supervision. We use the Mean Absolute Error ($M$), max F-measure ($F_m$), S-measure ($S_m$), and max E-measure ($E_m$) as evaluation metrics. (\textbf{Bold}: best; \underline{Underline}: second best.)}
\label{tab:abla}
\begin{tabular}{c c c cccccccccc}
 \hline 

\hline

\hline
\multirow{2}{*}{\#} & \multirow{2}{*}{Baseline} & \multirow{2}{*}{$GBA$}  & \multirow{2}{*}{$CDA$} & \multirow{2}{*}{$Skip$}   & \multirow{2}{*}{$EMI$} & \multirow{2}{*}{$\mathcal{L}_{ml}$} &$DEDA$    & $DCF$    &\multicolumn{2}{c}{DES}   & \multicolumn{2}{c}{STERE}\\
 \cline{10-11} \cline{12-13} 

& & & & &  & & \cite{zhao2020single} &  \cite{ji2021calibrated} & $M\downarrow$ &  $F_{\beta}\uparrow $   &  $M\downarrow$ &  $F_{\beta}\uparrow $  \\
\hline
1&\checkmark& & &  &  & & & &.018 & .941 &.038 &.917 \\
2&\checkmark&\checkmark & &  & & & & &.016 & .944 & .037 & .917  \\
3&\checkmark&\checkmark &\checkmark & & & & & &.016 & .946  & .036 & .919  \\
4&\checkmark&\checkmark &\checkmark & \checkmark & & & & &.015 & .947  & .036 & \bl{.923} \\
5&\checkmark&\checkmark &\checkmark & \checkmark &\checkmark & & & &\underline{.014} & \underline{.949} & \bl{.034} & \underline{.921}  \\
6&\checkmark&   & &\checkmark &\checkmark  & \checkmark  &\checkmark & & .016 & .946  &.041 & .914 \\
7&\checkmark&\checkmark  & &\checkmark &\checkmark  & \checkmark  & &\checkmark  & .017 & .946  &.037 & .918 \\
\rowcolor[RGB]{235,235,250} 

8&\checkmark&\checkmark &\checkmark & \checkmark &\checkmark &\checkmark & & & \bl{.013} & \bl{.952} &\underline{.035} & \underline{.921} \\

 \hline 

\hline

\hline
\end{tabular}
\end{center}
\end{table}

\label{abla:hida}
\subsection{Ablation study on Key Components} Tab. \ref{tab:abla} presents a thorough ablation study for each key component. We observe that by gradually adding proposed modules, our network leads to better performance. We also conduct experiments by replacing our proposed modules with several SOTA counterparts. Specifically, we compare our Granularity-Based Attention with the DEDA module proposed in \cite{zhao2020single}. Both our GBA and DEDA belong to the mask-guided attention modules. Specifically, DEDA leverages the depth map to dynamically learn the masked-guided attention map which is supervised by the ground truth. The learned attention map refers to the contrast to guide RGB learning. Differently, our mask is statically computed by the Otsu threshold by maximizing inter-class variance. The computed local regions refer to the fine-grained details which are further integrated with semantics cues. Empirically, by comparing ($\# 6-\#8$), our GBA performs favorably against DEDA, showing that our method can better leverage the depth cues to distinguish objects with different camera distances.  We also replace our encoder fusion (CDA) with the concurrent DCF \cite{ji2021calibrated} built upon channel attention. The main difference is that DCF is based on channel attention, while our CDA additionally leverages the spatial attention for better localization. By comparing ($\# 7-\#8$), we can observe that while CDA is replaced by the DCF, the performance drops significantly. This validates the effectiveness of our CDA with both channel and spatial attention.


\noindent\textbf{Design of Cross Dual Attention:} We verify in Tab. \ref{tab:ablacomp} the design of our encoder fusion by removing or replacing each component: ($C1$) Features are simply fused through addition; ($C2$) Features are fused through concatenation-convolution (\textbf{CC});  ($C3$) Features are firstly self-enhanced with vanilla CBAM before the addition fusion.  ($C4$) Features are firstly self-enhanced and later fused through \textbf{CC}. ($C5$) We explore the attention in a cross manner and fuse features with addition. We can observe the gain of attention modules by comparing ($C1 - C3 - C5$), the improvement from cross-domain interaction by comparing ($C3 - C5$), and the contribution of \textbf{CC} by comparing  ($C5-Ours$). These results validate the effectiveness of our proposed encoder fusion scheme.

\noindent\textbf{Design of Efficient Multi-Input Fusion:} We also verify the design of our decoder fusion in Tab. \ref{tab:ablacomp}: ($E1$) Features are fused with \textbf{CC}. ($E2$) Features are concatenated and fed into the ECA model before the convolution.  ($E3$) Features are fused with CC and then fed into the ECA. ($E4$) Based on the configuration $E2$, we further add a residual addition. By comparing ($E2 - E3$) and ($E4 -Ours$), we can observe that the ECA module performs better with a reduced channel size. The comparison between ($E2- E4$) validates the effectiveness of residual addition which propagates the hierarchical features.

\begin{table}
\footnotesize
\setlength\tabcolsep{8.5pt}
\setlength\extrarowheight{0pt}
\begin{center}
\caption{Ablation study on \textbf{encoder fusion} and \textbf{decoder fusion} designs. We use the Mean Absolute Error ($M$), max F-measure ($F_m$), S-measure ($S_m$), and max E-measure ($E_m$) as evaluation metrics. (\textbf{Bold}: best.)}
\label{tab:ablacomp}
\begin{tabular}[ht]{l l | c c | c c }
\hline

\hline

\hline
& \multirow{2}{*}{Configuration}  & \multicolumn{2}{c|}{DES} & \multicolumn{2}{c}{STERE} \\
& & \small $M\downarrow$ &  $F_{\beta}\uparrow $   &  $M\downarrow$ &  $F_{\beta}\uparrow $
\\
\hline
\rowcolor[RGB]{235,235,250} 

& \bl{HiDAnet} & \bl{.013} & \bl{.952} & \bl{.035} & \bl{.921} \\
\hline
C1& Add & .017 & .945 & .039 & .915 \\ 

C2& Cat + Conv & .016 & .946 & .039 & .916 \\

C3& Self + Add &  .015 & .948 & .036 &.918 \\

C4& Self + Cat + Conv &.014 & .949  &.037 & .917  \\

C5& Cross + Add &.015 & .947 & .036 &.919 \\

\hline
E1& Cat + Conv & .015 & .947 & .038 & .914\\ 

E2& Cat + ECA + Conv & .016 & .945  & .038 & .915   \\

E3& Cat + Conv + ECA &.015 &.949 & .037 & .916\\

E4& E2 + Residual & .014 & .950 & .036 & .920\\

\hline

\hline

\hline
\end{tabular}
\end{center}
\end{table}

\section{Conclusion}

In this paper, we propose an end-to-end HiDAnet for RGB-D saliency detection. Different from previous networks, we fully leverage fine-grained details and merge them with semantic cues through the local channel attention. Extensive evaluations on challenging RGB-D benchmarks indicate that our HiDAnet improves saliency detection in several challenging scenarios where the SOTA approaches fail, notably in cases where multiple objects with similar appearances but at distinct camera distances (granularity). Our method has the potential to be used in many other tasks, including semantic and instant segmentation.

\section*{Acknowledgements}
We gratefully acknowledge Zhuyun Zhou and Renato Martins for discussion and proofreading. 
This research is supported in part by the French National Research Agency through ANR CLARA (ANR-18-CE33-0004),  the French ADVANCES project ISITE-BFC project (ANR-15-IDEX-0003), and is financed by the French Conseil R\'egional de Bourgogne-Franche-Comt\'e.  

%
%
\bibliographystyle{IEEEtran}
\bibliography{egbib}

\vfill
\end{document}